\documentclass[10pt,twocolumn,letterpaper]{article}

\usepackage{cvpr}              

\usepackage{graphicx}
\usepackage{amsmath}
\usepackage{amssymb}
\usepackage{booktabs}

\usepackage{bm}
\usepackage{color}
\usepackage[accsupp]{axessibility}
\usepackage{makecell}
\usepackage[font=small,labelfont=bf,tableposition=top]{caption}
\usepackage{multirow}

\definecolor{modify}{rgb}{0.2,0.2,0.8}
\usepackage[pagebackref,breaklinks,colorlinks]{hyperref}

\usepackage[capitalize]{cleveref}
\crefname{section}{Sec.}{Secs.}
\Crefname{section}{Section}{Sections}
\Crefname{table}{Table}{Tables}
\crefname{table}{Tab.}{Tabs.}

\def\cvprPaperID{5780} 
\def\confName{CVPR}
\def\confYear{2022}

\begin{document}

\title{Deep Hyperspectral-Depth Reconstruction Using Single Color-Dot Projection}

\author{Chunyu Li, Yusuke Monno, and Masatoshi Okutomi\\
Tokyo Institute of Technology, Tokyo, Japan\\
{\tt\small \{lchunyu,ymonno\}@ok.sc.e.titech.ac.jp, mxo@ctrl.titech.ac.jp}
}
\maketitle


\begin{abstract}
Depth reconstruction and hyperspectral reflectance reconstruction are two active research topics in computer vision and image processing. Conventionally, these two topics have been studied separately using independent imaging setups and there is no existing method which can acquire depth and spectral reflectance simultaneously in one shot without using special hardware. In this paper, we propose a novel single-shot hyperspectral-depth reconstruction method using an off-the-shelf RGB camera and projector. Our method is based on a single color-dot projection, which simultaneously acts as structured light for depth reconstruction and spatially-varying color illuminations for hyperspectral reflectance reconstruction. To jointly reconstruct the depth and the hyperspectral reflectance from a single color-dot image, we propose a novel end-to-end network architecture that effectively incorporates a geometric color-dot pattern loss and a photometric hyperspectral reflectance loss. Through the experiments, we demonstrate that our hyperspectral-depth reconstruction method outperforms the combination of an existing state-of-the-art single-shot hyperspectral reflectance reconstruction method and depth reconstruction method.

\end{abstract}

\section{Introduction}
\label{sec:intro}

\begin{figure}[!tbp]
  \centering
 \includegraphics[width=\hsize]{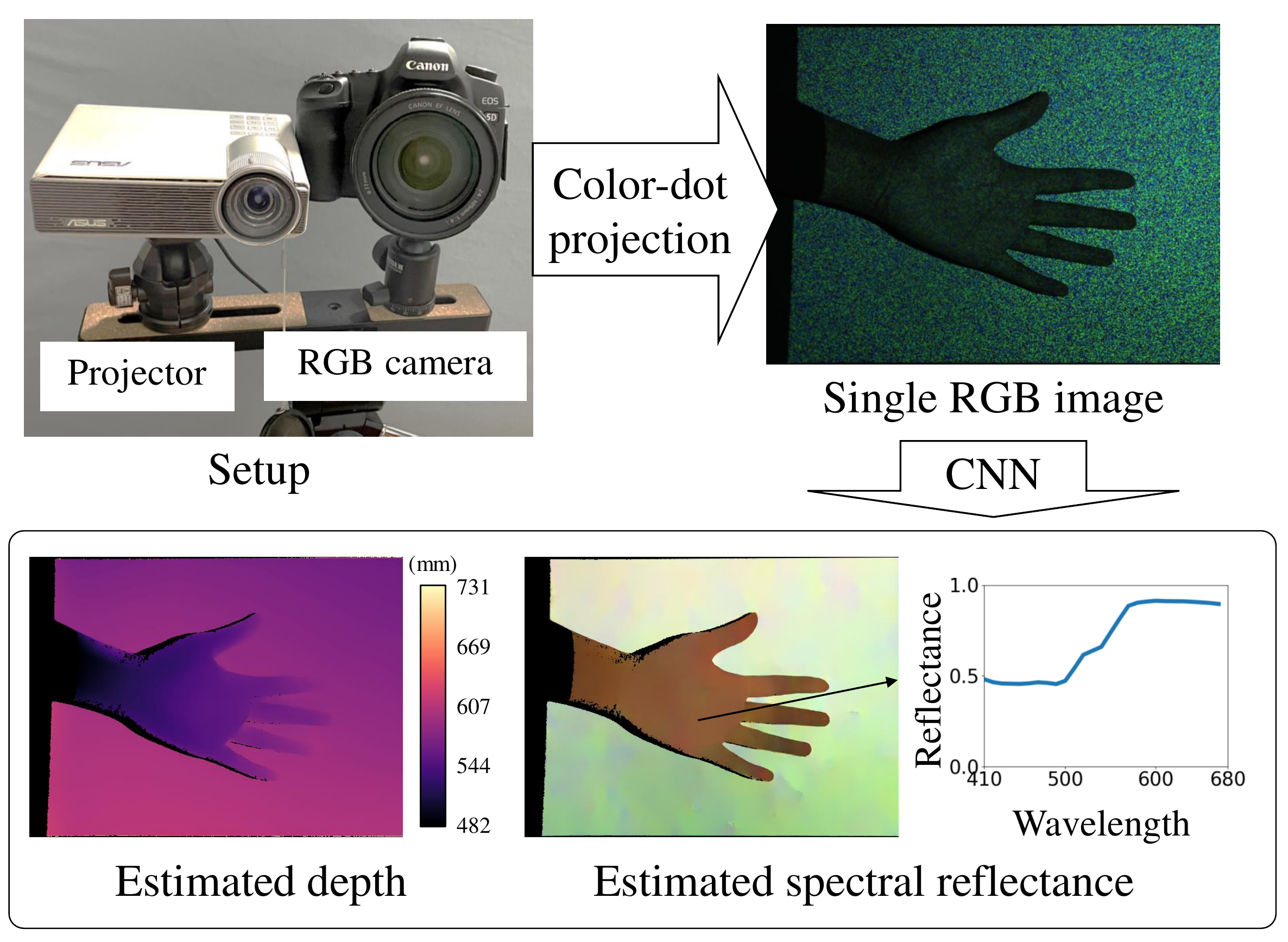}
 \vspace{-6mm}
  \caption{The overview of our system. From a single RGB image captured with a random color-dot projection, 
  we simultaneously reconstruct the depth and the spectral reflectance for each pixel.}
  \label{fig:introduction}
      \vspace{-2.5mm}
\end{figure}

Depth reconstruction and hyperspectral reflectance reconstruction (spectral reconstruction, for short) are two active research areas in the fields of computer vision and image processing. Depth reconstruction aims at obtaining a scene's depth map, which presents the distances from the camera to each scene point. On the other hand, spectral reconstruction aims at acquiring scene's spectral reflectance information, which provides the wavelength-by-wavelength reflectance of each scene point. Since the depth and the spectral reflectance provides the scene's geometric and photometric properties, respectively, simultaneously acquiring them, which we refer to as hyperspectral-depth reconstruction, has various potential applications such as cultural heritage~\cite{chane2013integration,kim2014hyper3d}, artwork authentication~\cite{polak2017hyperspectral}, material classification~\cite{brusco2006system,liang2014remote}, plant modeling~\cite{liang20133d}, and relighting~\cite{wilkie2002tone}.


Although depth reconstruction and spectral reconstruction have been studied separately, some systems are recently designed to simultaneously acquire both the depth and the spectral reflectance. They typically combine a conventional depth-sensing technology with a hyperspectral camera~\cite{heist20185d,xiong2017snapshot,zhu2018hyperspectral,rueda2019snapshot,diaz2018hyperspectral,wang2016simultaneous}. However, the requirement of a hyperspectral camera makes the system high cost.
Some other systems use a standard RGB camera in conjunction with a variable and controllable light source, which emits temporally-changing illuminations to acquire multi-band spectral observations ~\cite{li2019pro,li2021spectral,xu2020hyperspectral,kitahara2015simultaneous,nam2014multispectral,ozawa2017hyperspectral,hirai2016measuring}. 
However, these systems require multiple shots and thus are not applicable to dynamic scenes. 
Very recently, Baek et al. have proposed a single-shot system that uses a standard RGB camera and a diffractive optical element attached in front of the camera~\cite{baek2021single}. Although this system realizes a compact design using existing optical components, it still requires customized hardware design.


In this paper, we propose a novel single-shot system to simultaneously acquire the depth and the spectral reflectance using a standard RGB camera and an off-the-shelf RGB projector (see Fig.~\ref{fig:introduction}). Our system is based on a single random color-dot projection, which simultaneously acts as structured light for depth reconstruction and spatially-varying color illuminations for spectral reconstruction. Since the random color dots provide a unique code pattern and three distinct RGB color illuminations for each local region, we exploit these cues for the hyperspectral-depth reconstruction.
To effectively reconstruct the depth and the spectral reflectance from a single color-dot image, we propose a novel end-to-end deep learning method. Since the location of an observed color-dot pattern depends on the scene depth, we perform the joint learning of the depth and the spectral reflectance to improve the accuracy of each other, by considering the geometric warping of the color-dot pattern. Furthermore, to address the difficulty of constructing a real-world hyperspectral-depth dataset, we develop a spectral renderer to generate a synthetic dataset using a spectral rendering model under the color-dot illumination. 
Main contributions of this work are summarized as follows.
\begin{enumerate}
  \vspace{-1mm}
  \item We propose the first single-shot hyperspectral-depth reconstruction system using a standard RGB camera and an off-the-shelf RGB projector without any hardware modifications.
  \vspace{-1mm}
  \item We propose a novel network architecture and end-to-end learning method using a spectral renderer to simultaneously reconstruct the depth and the spectral reflectance from a single color-dot pattern image.
  \vspace{-1mm}
  \item We experimentally validate the effectiveness of our system for synthetic and real-world scenes.
\end{enumerate}

\section{Related Works}

Existing systems for hyperspectral-depth reconstruction are roughly classified into hyperspectral camera-based  systems~\cite{xiong2017snapshot,zhu2018hyperspectral,rueda2019snapshot,diaz2018hyperspectral,wang2016simultaneous,heist20185d} and controllable lighting-based systems~\cite{li2019pro,li2021spectral,xu2020hyperspectral,kitahara2015simultaneous,nam2014multispectral,ozawa2017hyperspectral,hirai2016measuring}.

Most of the hyperspectral camera-based systems acquire the depth and the spectral reflectance data by replacing the RGB camera of an existing depth-sensing technology, such as structured light~\cite{heist20185d,diaz2018hyperspectral}, ToF~\cite{rueda2019snapshot}, stereo~\cite{wang2016simultaneous}, and light fields~\cite{xiong2017snapshot,zhu2018hyperspectral}, with a hyperspectral camera. Although these systems can realize the single-shot acquisition of the depth and the spectral reflectance, the necessity of a hyperspectral camera brings high cost. Also, the integration of a hyperspectral camera into a depth-sensing device requires highly complicated and dedicated hardware design.

Controllable lighting-based systems are based on a traditional 3D reconstruction method that uses extra light sources, such as structured light~\cite{li2019pro,xu2020hyperspectral,hirai2016measuring} and photometric stereo~\cite{li2021spectral,kitahara2015simultaneous,nam2014multispectral,ozawa2017hyperspectral}. These systems use a standard RGB camera and observe spectral measurements by temporally changing illumination spectrum. However, since these systems require multiple shots, they are limited to static scenes.

There are two other classes of closely related methods: lighting-based hyperspectral imaging methods using an RGB camera~\cite{Cui,Han2,Park,Hidaka} and  
deep-learning-based active stereo methods~\cite{riegler2019connecting,zhang2018activestereonet,baek2021polka,fanello2016hyperdepth} (especially, Connecting the Dots~\cite{riegler2019connecting}, which learns to reconstruct the depth from a single gray-scale-dot pattern image, is the closest work to ours).
Although these methods inspired us, they only reconstruct either the depth or the spectral reflectance. In contrast, our method simultaneously reconstructs the depth and the spectral reflectance from a single color-dot image based on end-to-end network learning, which consequently enables us to improve the accuracy of each other.

\section{Proposed Method}

\begin{figure}[!tbp]
  \centering
 \includegraphics[width=\hsize]{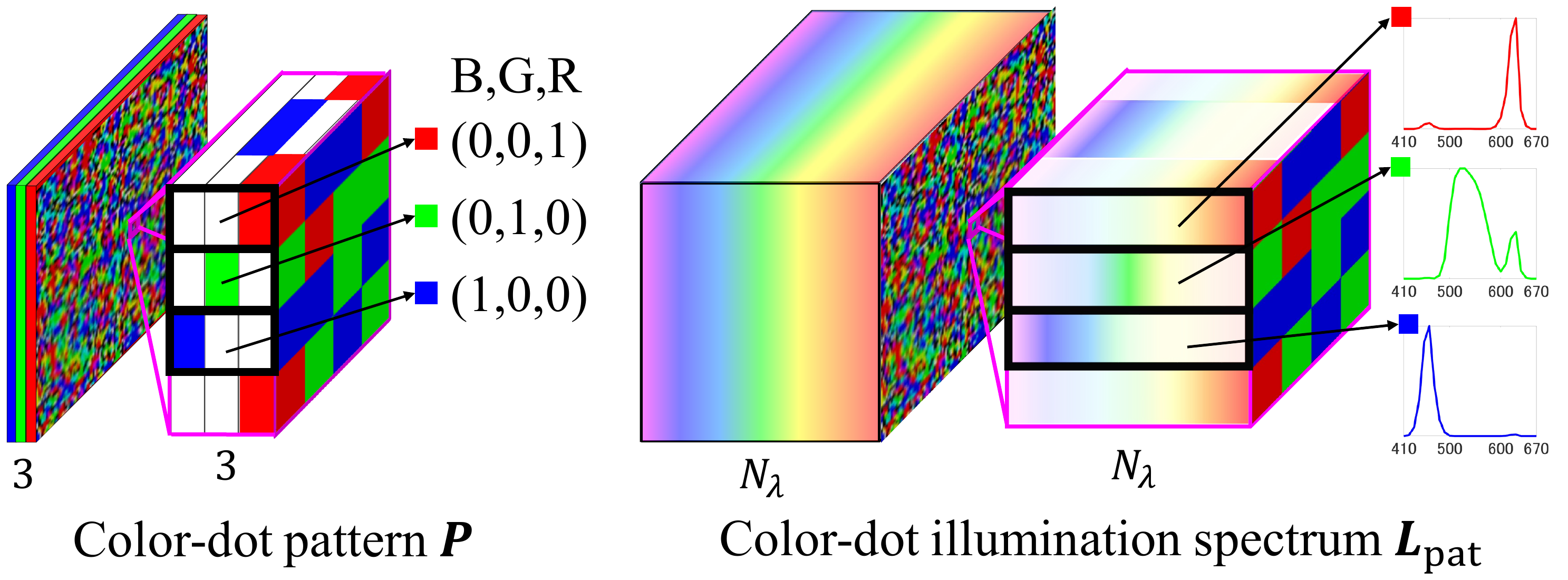}
 \vspace{-6mm}
  \caption{Color-dot representations. Left: Color-dot pattern $\bm{P}$, which is generated by randomly filling each projector pixel with one of three binary codes: R~(0,0,1), G~(0,1,0), and B~(1,0,0). Right: Color-dot illumination spectrum $\bm{L}_{\rm pat}$, which has $N_{\lambda}$-dimensional illumination spectrum at each pixel.}
  \label{fig:pattern}
      \vspace{-2mm}
\end{figure}

\begin{figure*}[!tbp]
  \centering
 \includegraphics[width=\hsize]{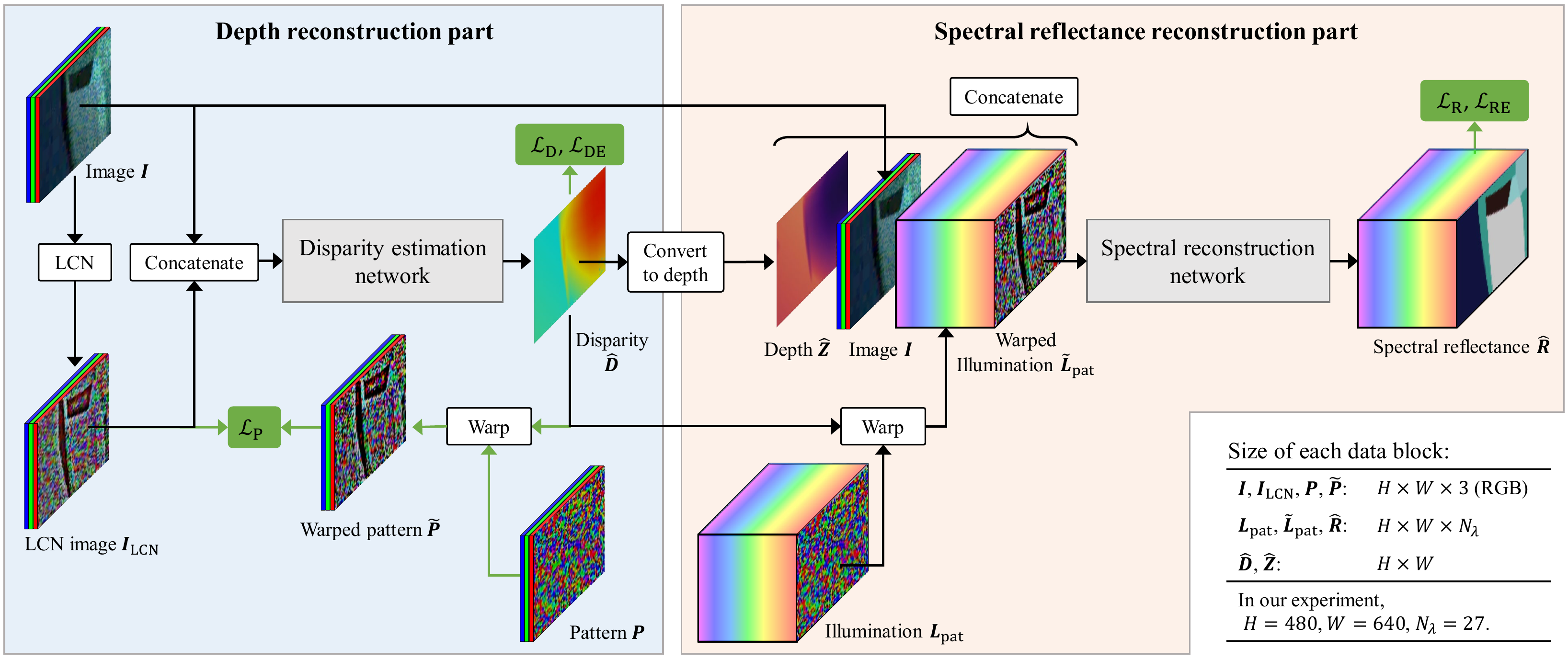}
 \vspace{-6mm}
  \caption{The overview of our end-to-end network architecture. As the first part, the disparity estimation network estimates the disparity map $\bm{\hat{D}}$ from the captured image $\bm{I}$ and the local contrast normalization~(LCN) image $\bm{I}_{\rm LCN}$. Then, the estimated disparity map $\bm{\hat{D}}$ is converted to the depth map $\bm{\hat{Z}}$. As the second part, the spectral reconstruction network estimates the spectral reflectance image $\bm{\hat{R}}$ from the inputs of the captured image $\bm{I}$, the estimated depth map $\bm{\hat{Z}}$, and the warped illumination spectrum $\tilde{\bm{L}}_{\rm pat}$.
  The two networks are trained in an end-to-end manner using both geometric losses ($\mathcal{L}_{\rm D}$, $\mathcal{L}_{\rm DE}$ and $\mathcal{L}_{\rm P}$) and photometric losses ($\mathcal{L}_{\rm R}$ and $\mathcal{L}_{\rm RE}$).}
  \label{fig:network}
      \vspace{-1mm}
\end{figure*}

\subsection{Random Color-Dot Projection}
\label{ssec:colorpattern}

In our system, we use an off-the-shelf RGB projector and a standard RGB camera to capture a single color-dot pattern image. The extrinsic and intrinsic parameters of the projector-camera system are pre-calibrated. The spectral sensitivity of the RGB camera and the spectral power distributions of the projector's RGB primaries are assumed to be known or pre-estimated. 

As shown in the left figure of Fig.~\ref{fig:pattern}, the projector is used to project a single color-dot pattern to acquire geometric and spectral observations. The color-dot pattern $\bm{P}$ is generated by randomly filling each projector pixel with one of the three code words representing the projector's RGB primaries: R~(0,0,1), G~(0,1,0), and B~(1,0,0).

As the geometric observation, the random pattern provides a locally unique code for establishing correspondences between the captured color-dot image and the reference color-dot pattern $\bm{P}$. As the spectral observation, the projector's RGB primaries provide three distinct illuminations, which results in the information from nine spectral bands (i.e., 3 illuminations $\times$ 3 color channels) by assuming locally uniform spectral reflectance. The illumination spectrum representation of the projected color-dot pattern is denoted by $\bm{L}_{\rm pat}$ and shown in the right figure of Fig.~\ref{fig:pattern}, where $N_{\lambda}$ denotes the dimension discretized from the continuous wavelength domain (specifically, we used the sampling of every 10nm from 410nm to 670nm, i.e., $N_\lambda$=27).




\subsection{End-to-End Network Architecture}


Figure~\ref{fig:network} illustrates the overview of our end-to-end network architecture. Based on the geometric and the photometric cues that can be observed from the color-dot projection, we reconstruct a disparity map $\bm{\hat{D}}$ and a spectral reflectance image $\bm{\hat{R}}$ from a single color-dot image $\bm{I}$, where the image $\bm{I}$ is rectified in advance using the extrinsic and intrinsic parameters of the projector-camera system. For the reconstruction, we apply two deep convolutional neural networks: disparity estimation network and spectral reconstruction network.

Firstly, the disparity estimation network estimates the disparity map $\bm{\hat{D}}$ with the inputs of the captured image $\bm{I}$ and the local contrast normalization~(LCN) image $\bm{I}_{\rm LCN}$. LCN is applied to extract the color-dot pattern from $\bm{I}$. The estimated disparity is then converted to the depth map $\bm{\hat{Z}}$ using the calibrated parameters of the projector-camera system.

Then, the spectral reconstruction network estimates the spectral reflectance image $\bm{\hat{R}}$ with the inputs of the captured image $\bm{I}$, the estimated depth map $\bm{\hat{Z}}$, and the warped illumination spectrum $\tilde{\bm{L}}_{\rm pat}$. In this study, our aim is to reconstruct the spectral reflectance, which is inherent to a target object and irrelevant to the illumination and the scene geometry such as the shading. Since the depth provides an important cue to eliminate the effect of the shading from the estimated spectral reflectance, we input the depth map into the spectral reconstruction network. In addition, to provide the correct illumination information for each camera pixel,  we input the illumination spectrum $\tilde{\bm{L}}_{\rm pat}$, which can be generated by warping the color-dot illumination spectrum $\bm{L}_{pat}$ from the projector viewpoint to the camera viewpoint based on the estimated disparity map.


The two networks are trained in a supervised and end-to-end manner using both geometric losses ($\mathcal{L}_{\rm D}$, $\mathcal{L}_{\rm DE}$ and $\mathcal{L}_{\rm P}$) and photometric losses ($\mathcal{L}_{\rm R}$ and $\mathcal{L}_{\rm RE}$). 
In our training process, the error of the estimated disparity will lead to wrong shading inference and wrong illumination warping for the spectral reconstruction network, meaning that the accuracy of the disparity affects the accuracy of the spectral reflectance. Thus, jointly training the two networks contributes to the improvement of the accuracy for both the disparity and the spectral reflectance, as we will demonstrate in the experimental result section.

\subsubsection{Disparity Estimation Network}
\label{sssec:disparitynetwork}

The disparity estimation network produces an output image of the same resolution as the input with left-right disparity information.
Since the appearance of the color-dot pattern in the captured image $\bm{I}$ depends on various spatially-varying factors such as the shading and the texture, we pre-process the captured image to extract the projected color-dot pattern by applying LCN~\cite{jarrett2009best, zhang2018activestereonet, riegler2019connecting}.
Following~\cite{riegler2019connecting}, for each pixel $(u,v)$ and each color channel $n$, we compute the local mean $\mu$ and the standard deviation $\sigma$ of a small local region (11 $\times$ 11 in our experiments) centered at pixel coordinate $(u,v)$. These local statistics are used to normalize the current pixel intensity as
\begin{equation}
\bm{I}_{\rm LCN}(u,v,n)=\frac{\bm{I}(u,v,n)-\bm{\mu}(u,v,n)}{\bm{\sigma}(u,v,n)+\eta},
\label{eq:lcn}
\end{equation}
where $\eta$ is a small constant to avoid numerical instabilities.


Then, we concatenate the LCN image $\bm{I}_{\rm LCN}$ with the original image $\bm{I}$ to form a six-channel input for the disparity estimation network, where the disparity is defined by the x-coordinate difference of the corresponding pixels between the captured image and the reference color-dot pattern. Given the estimated disparity map $\bm{\hat{D}}$, the scene depth $\bm{\hat{Z}}$ can be calculated as
\begin{equation}
\bm{\hat{Z}}(u,v) = bf / \bm{\hat{D}}(u,v),
\label{eq:depth}
\end{equation}
where $b$ is the baseline of the projector-camera system and $f$ is the focal length of the rectified camera.

We design the network architecture based on the Disparity Decoder presented in~\cite{riegler2019connecting}.
This network consists of a contractive part and an expanding part with long-range links between them. 
In total, the network has 32 convolution layers and each of them is followed by ReLU. The final layer is followed by a scaled sigmoid non-linearity which constrains the output disparity map to the range between 0 and the maximum of the disparity. The network details can be found in the supplementary document.


\subsubsection{Spectral Reconstruction Network}

We next estimate the spectral reflectance image $\bm{\hat{R}}$ using the spectral reconstruction network. The captured image $\bm{I}$, the predicted scene depth $\bm{\hat{Z}}$ and the warped illumination spectrum to the camera viewpoint $\tilde{\bm{L}}_{\rm pat}$ are concatenated and passed to the spectral reconstruction network.
As the disparity map provides the pixel correspondences between the captured image and the reference color-dot pattern, the warped illumination spectrum $\tilde{\bm{L}}_{\rm pat}$ can be calculated as 
\begin{equation}
\tilde{\bm{L}}_{\rm pat}(u,v,\lambda)=\bm{L}_{\rm pat}(u-\bm{\hat{D}}(u,v),v,\lambda).
\label{eq:warppattern}
\end{equation}
Since the disparity is estimated  with sub-pixel accuracy, we apply bilinear interpolation for the resampling of the warped illumination spectrum.

The network architecture of the spectral reconstruction network is similar to that of the disparity estimation network with the difference of the input and the output channels.
The range of the output spectral reflectance is constrained between 0 and 1 by the scaled sigmoid non-linearity.

\subsection{Loss Function}
The loss function for end-to-end training is described as
\begin{equation}
\mathcal{L}=\sum_{(u,v) \in \mathcal{V}}\mathcal{L}_{\rm D} + \omega_{\rm DE}\mathcal{L}_{\rm DE} + \omega_{\rm P}\mathcal{L}_{\rm P} + \omega_{\rm R}\mathcal{L}_{\rm R} + \omega_{\rm RE}\mathcal{L}_{\rm RE},
\label{eq:loss}
\end{equation}
including geometric losses (disparity loss $\mathcal{L}_{\rm D}$, disparity edge loss $\mathcal{L}_{\rm DE}$, and pattern loss $\mathcal{L}_{\rm P}$) and photometric losses (spectral reflectance loss $\mathcal{L}_{\rm R}$ and spectral reflectance edge loss $\mathcal{L}_{\rm RE}$). The balance of each loss is determined by the parameters $\omega_{\rm DE}$, $\omega_{\rm P}$, $\omega_{\rm R}$, and $\omega_{\rm RE}$.
As the cast shadows that are apparent in the input image are meaningless in the network training, we binarize the input image to mask out the shadows and calculate the losses only for the non-shadow pixel set $(u,v) \in \mathcal{V}$.

Disparity loss $\mathcal{L}_{\rm D}$ and spectral reflectance loss $\mathcal{L}_{\rm R}$ compute the mean squared error between the ground truth and the estimated value as
\begin{equation}
\begin{split}
\mathcal{L}_{\rm D}&=\|{\bm{\hat{D}}(u,v)-\bm{D}_{\rm gt}(u,v)}\|^2,\\
\mathcal{L}_{\rm R}&=\sum_{\lambda}\|{\bm{\hat{R}}(u,v,\lambda)-\bm{R}_{\rm gt}(u,v,\lambda)}\|^2,
\label{eq:suploss}
\end{split}
\end{equation}
where $\bm{D}_{\rm gt}$ is the ground-truth disparity and
$\bm{R}_{\rm gt}$ is the ground-truth spectral reflectance.

For pattern loss $\mathcal{L}_{\rm P}$, we take the advantage of the projector-camera setup to strengthen geometric constraints. To this end, we warp the reference color-dot pattern $\bm{P}$ to the camera viewpoint using the estimated disparity $\bm{\hat{D}}$ as
\begin{equation}
\tilde{\bm{P}}(u,v)=\bm{P}(u-\bm{\hat{D}}(u,v),v),
\label{eq:warpillumination}
\end{equation}
where $\tilde{\bm{P}}$ is the warped pattern. Then, we calculate the loss between the LCN image $\bm{I}_{\rm LCN}$ and the warped color-dot pattern $\tilde{\bm{P}}$ as
\begin{equation}
\mathcal{L}_{\rm P}=\|{\bm{I}_{\rm LCN}(u,v)-\tilde{\bm{P}}(u,v)}\|_C,
\label{eq:patloss}
\end{equation}
where $\|\cdot\|_C$ denotes the smooth Census transform~\cite{hafner2013census}.

As the color-dot pattern is relatively sparse, we further add disparity edge loss $\mathcal{L}_{\rm DE}$ and spectral reflectance edge loss $\mathcal{L}_{\rm RE}$ for predicting accurate and sharp boundaries.
To this end, we use Sobel operator~\cite{sobel} to perform 2D spatial gradient calculation on the disparity and the spectral reflectance to enhance the boundaries.
We apply a pair of Sobel convolution kernels to produce the approximate gradients of each pixel in the vertical and horizontal directions. Then, we calculate the errors of vertical gradients and horizontal gradients separately and add then up together.
We formulate the losses $\mathcal{L}_{\rm DE}$ and $\mathcal{L}_{\rm RE}$ as
\begin{equation}
\begin{split}
\mathcal{L}_{\rm DE}=&\|{\bm{\hat{D}}^{\rm V}(u,v)-\bm{D}^{\rm V}_{\rm gt}(u,v)}\|^2 \\
&+ \|{\bm{\hat{D}}^{\rm H}(u,v)-\bm{D}^{\rm H}_{\rm gt}(u,v)}\|^2,
\label{eq:dispedgeloss}
\end{split}
\end{equation}
\begin{equation}
\begin{split}
\mathcal{L}_{\rm RE}=&\sum_{\lambda} \|{\bm{\hat{R}}^{\rm V}(u,v,\lambda)-\bm{R}^{\rm V}_{\rm gt}(u,v,\lambda)}\|^2 \\
&+ \|{\bm{\hat{R}}^{\rm H}(u,v,\lambda)-\bm{R}^{\rm H}_{\rm gt}(u,v,\lambda)}\|^2,
\label{eq:refedgeloss}
\end{split}
\end{equation}
where the values with superscripts ${\rm V}$ and ${\rm H}$ denote the vertical gradient and the horizontal gradient, respectively.

\subsection{Hyperspectral-Depth Dataset Generation}
\label{ssec:cgdata}

Since it is difficult to simultaneously acquire accurate depth and spectral reflectance as a large-scale ground-truth dataset in real-world situations, we developed a spectral renderer to generate a synthetic dataset with rendered RGB color-dot images, ground-truth disparity maps, and ground-truth spectral reflectance images by extending the algorithm of a structured-light renderer~\cite{riegler2019connecting}.

We render the scene with randomly populated 3D models using spectral reflectance samples. For simplicity, we first obtain the corresponding 3D point $\bm{x}$ for each pixel by computing the intersection of the camera ray and the 3D surface, and then acquire the ground-truth depth value as the z-coordinate of the 3D point in the camera coordinate system. The ground-truth spectral reflectance $\bm{r}$ of this 3D point is also obtained.
According to Eq.~(\ref{eq:depth}), we can obtain the ground-truth disparity from the depth value. The illumination spectrum $\bm{l}$ for the 3D point is determined by the corresponding pattern code which can be obtained by reprojecting the 3D point to the projector's image plane. 

Suppose that the camera response is linear and inter-reflection and ambient illumination are negligible, the camera’s pixel intensity $I$ of $n$-th color channel is calculated based on the spectral rendering model~\cite{li2019pro} as
\begin{equation}
I(n) =s\int_{\Omega_\lambda}c(n,\lambda)l(\lambda)r(\lambda)d\lambda,
\label{eq:rendering}
\end{equation}
where $c(n,\lambda)$ is the $n$-th channel camera spectral sensitivity and $\lambda$ represents the wavelength. $\Omega_\lambda$ is the wavelength range that the projector emits the illumination (410nm to 670nm for our used projector).
$s$ is the shading factor, which describes the proportion of the reflected radiance leaving the surface point $\bm{x}$ with respect to the intensity of the emitted light from the projector at position $\bm{x}_{\rm pro}$. Assuming that the 3D point has Lambertian reflectance and the projected illumination follows the inverse-square law, we define the shading factor $s$ as
\begin{equation}
s=\frac{1}{\|\bm{x}_{\rm pro}-\bm{x}\|^2}\times\frac{\bm{x}_{\rm pro}-\bm{x}}{\|\bm{x}_{\rm pro}-\bm{x}\|}\cdot \bm{n},
\label{eq:shading}
\end{equation}
where $\bm{n}$ is the normal of the point $\bm{x}$. The first term represents the quadratic attenuation with respect to the distance of the object point $\bm{x}$ from the projector $\bm{x}_{\rm pro}$. The second term represents the inner product of the normalized lighting vector and the point normal $\bm{n}$.
In practice, the continuous wavelength domain $\Omega_\lambda$ is discretized to $N_\lambda$ dimension (we sampled at every 10nm from 410nm to 670nm, i.e., $N_\lambda$=27). The observed RGB intensity $[I(R), I(G), I(B)]^T$ can be computed by the matrix form as
\begin{equation}
\left[\begin{matrix} I(R) \\ I(G) \\ I(B) \end{matrix}\right] = s \bm{c}^T {\rm Diag}(\bm{l}) \bm{r},
\label{eq:renderingmatrixform}
\end{equation}
where $\bm{r} \in \mathbb{R}^{N_\lambda}$ represents the spectral reflectance, $\bm{l} \in \mathbb{R}^{N_\lambda}$ is the illumination spectrum corresponding one of the projector's RGB primaries, $\bm{c} \in \mathbb{R}^{N_\lambda \times 3}$ is the camera sensitivity matrix, and ${\rm Diag(\cdot)}$ is a square diagonal matrix function.

We used the same camera spectral sensitivity, projector illumination spectrum, and geometrically calibration parameters as our actual setup, which is described in Sec\ref{ssec:setup}.

\begin{figure}[!tbp]
  \centering
  \begin{minipage}[t]{0.29\hsize}
      \centering
        \includegraphics[width=\hsize]{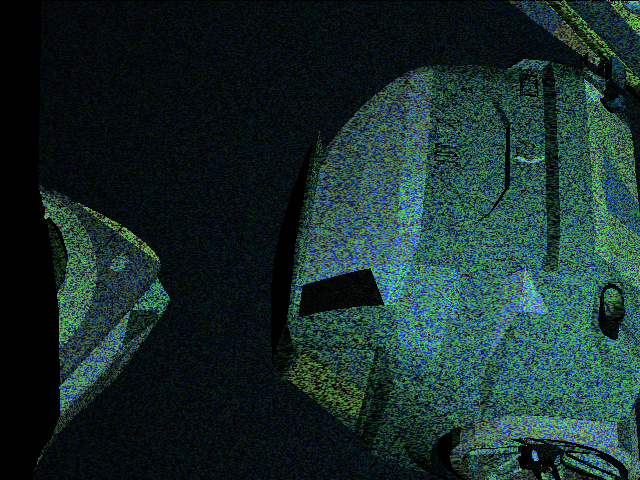}\\
        \footnotesize{(a) Color-dot image}
     \end{minipage}
    \begin{minipage}[t]{0.375\hsize}
      \centering
        \includegraphics[width=0.77\hsize]{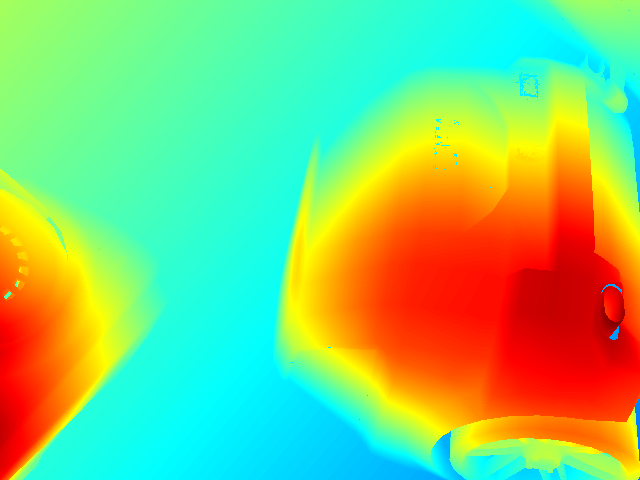}
        \includegraphics[width=0.14\hsize]{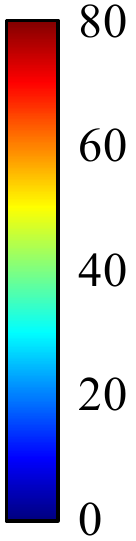}\\
        \footnotesize{(b) Disparity map}
     \end{minipage}
    \begin{minipage}[t]{0.29\hsize}
      \centering
        \includegraphics[width=\hsize]{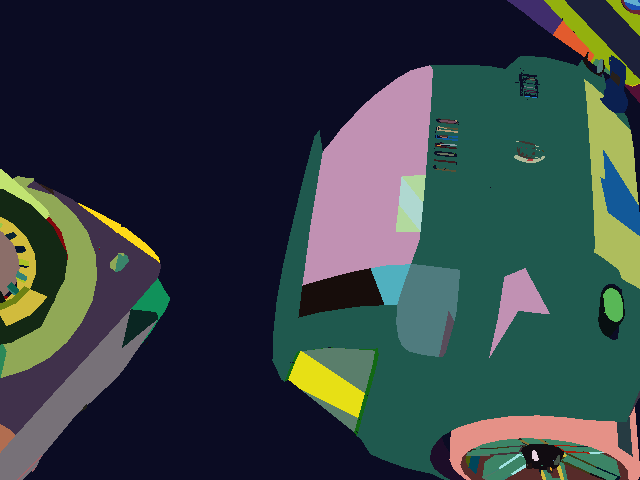}\\
        \footnotesize{(c) sRGB}
     \end{minipage}
     \vspace{-2mm}
  \caption{Spectral rendering examples: (a) A rendered input RGB image with the projected color-dot pattern. (b) A ground-truth disparity map, and (c) An sRGB image converted from the ground-truth spectral reflectances.}
  \label{fig:renderingexample}
     \vspace{-2mm}
\end{figure}

\begin{table*}[t]
  \small
\begin{minipage}{0.31\hsize}
 \caption{Evaluation metrics.}
  \vspace{-2mm}
 \centering
  \begin{tabular}{p{0.91\hsize}}
   \toprule
 \textbf{Depth:}\\
 ${\rm RMSE}=\sqrt{{\rm mean} \left[ \left(\bm{\hat{Z}}-\bm{Z}_{\rm gt}\right)^2\right]}$\\
 $\theta_i=\%$ of $\bm{\hat{Z}}(u,v)$ subject to $\max\left(\frac{\bm{\hat{Z}}(u,v)}{\bm{Z}_{\rm gt}(u,v)},\frac{\bm{Z}_{\rm gt}(u,v)}{\bm{\hat{Z}}(u,v)}\right) < 1.03^i$\\
 \midrule[0.2pt]
 \textbf{Spectral reflectance:}\\
 ${\rm RMSE}=\sqrt{{\rm mean}\left[ \left( \bm{\hat{R}}-\bm{R}_{\rm gt}\right)^2\right]}$\\
 ${\rm MRAE}={\rm mean}\left(|\bm{\hat{R}}-\bm{R}_{\rm gt}|/\bm{R}_{\rm gt}\right)$\\
   \bottomrule
  \end{tabular}
 \label{table:evaluationmetrics}
 \end{minipage}
 \hspace{0.3em}
\begin{minipage}{0.65\hsize}
 \caption{Quantitative comparison with the state-of-the-art methods on all the test scenes.}
  \vspace{-2mm}
 \centering
  \small
  \begin{tabular}{lcccccc}
   \toprule
   &\multicolumn{4}{c}{Depth}&\multicolumn{2}{c}{Spectral reflectance}\\
\cmidrule(lr){2-5}
\cmidrule(lr){6-7}
    & $\theta_1$ $\uparrow$ & $\theta_2$ $\uparrow$ & $\theta_3$ $\uparrow$ & RMSE $\downarrow$& MRAE $\downarrow$ & RMSE ($\times10^{-2}$) $\downarrow$    \\
   \midrule          
   AdaBins~\cite{bhat2021adabins} &53.00 & 82.61 & 93.52 & 24.60 &-&- \\
   Connecting~\cite{riegler2019connecting} & 98.02 & 98.72 & 99.11 & 8.83  &-&-\\
   Basis~\cite{Han2}&-&-&-&-& 0.38& 8.02  \\
   AWAN~\cite{li2020adaptive}&-&-&-&-& 0.34 & 7.93   \\
   Ours & \textbf{98.18} & \textbf{99.17} & \textbf{99.58} & \textbf{6.10} & \textbf{0.32} & \textbf{5.31} \\
   \bottomrule
  \end{tabular}
 \label{table:stateoftheart}
 \end{minipage}
      \vspace{-2mm}
\end{table*}

\vspace{-1mm}
\section{Experimental Results}

\subsection{Setup and Implementation Details}
\label{ssec:setup}

We used an ASUS P3B projector and Canon EOS 5D Mark-II digital camera for our projector-camera system. The spectral power distributions of the projector's RGB primaries were measured by using a StellarNet BlueWave-VIS Spectrometer and they are shown in Fig.~\ref{fig:pattern}.
The spectral sensitivity of EOS 5D camera was obtained from the public database of~\cite{Jiang}. To calibrate the projector-camera system geometrically, we used the calibration method of~\cite{shahpaski2017simultaneous}.

For the synthetic dataset generation described in Sec.\ref{ssec:cgdata}, we used ShapeNet Core dataset~\cite{chang2015shapenet} as the 3D models. We randomly placed the models at a distance from 0.3m to 1m and then randomly assigned the ground-truth spectral reflectance data to different texture parts of each 3D model.
We generated 8,192 scenes for training, and 256 scenes for testing. For the training data, we used chair and car models in ShapeNet Core and the spectral reflectance data of 1,269 Munsell color chips~\cite{Munsell}. For the testing data, we used camera, airplane, and watercraft models in ShapeNet Core and the spectral reflectance data of a standard X-Rite colorchart with 24 patches, which are unseen in the training data.
The images were rendered with the resolution of 640$\times$480.
Rendering examples of an RGB color-dot image, a ground-truth disparity map, and an sRGB color image converted from the ground-truth spectral reflectances are shown in Fig.~\ref{fig:renderingexample}.

We implemented the proposed method in PyTorch and trained our model using Adam optimizer~\cite{kingma2014adam}. The learning rate was set as $1.0\times 10^{-4}$.
The loss weights in Eq.~(\ref{eq:loss}) were empirically set as $\omega_{\rm DE}=100$, $\omega_{\rm P}=0.2$, $\omega_{\rm R}=1$, and $\omega_{\rm RE}=8$.
We used full-size 640x480 images for training. The total number of training epochs is 200 with batch size of 8.
Training our model takes around 57 hours in total with one NVIDIA GeForce RTX 2080 Ti 11G GPU. 

\subsection{Evaluation on Synthetic Data}
\label{ssec:simulation}
We first qualitatively and quantitatively evaluate our proposed method on the test set of the synthetic dataset generated using the spectral renderer.

For the depth evaluation, we use root mean squared errors (RMSE) and threshold accuracy ($\theta_i$) used in~\cite{bhat2021adabins}.
For the spectral reflectance evaluation, we use RMSE and
mean relative absolute error (MRAE) used in~\cite{li2020adaptive}.
These metrics are formulated in Table~\ref{table:evaluationmetrics}, where ${\rm mean}(\cdot)$ computes the arithmetic mean.

\begin{figure}[t]
    \centering
    \begin{minipage}{\hsize}
    \centering
      \begin{minipage}[t]{0.32\hsize}
          \centering
            \includegraphics[width=\hsize]{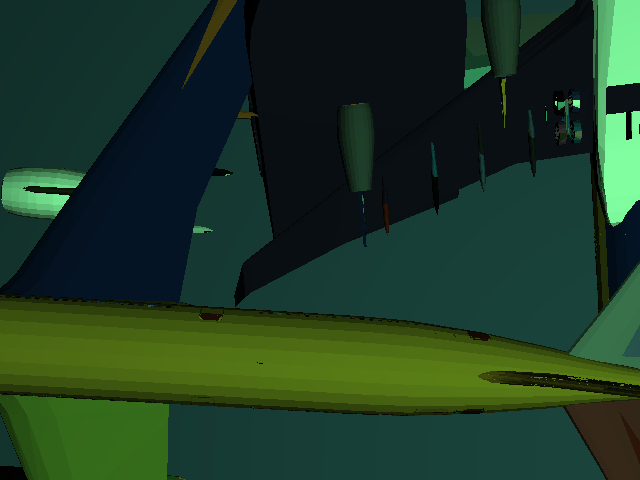}\\
            \scriptsize{Input for AdaBins, Basis, and AWAN}
         \end{minipage}
        \begin{minipage}[t]{0.32\hsize}
          \centering
            \includegraphics[width=\hsize]{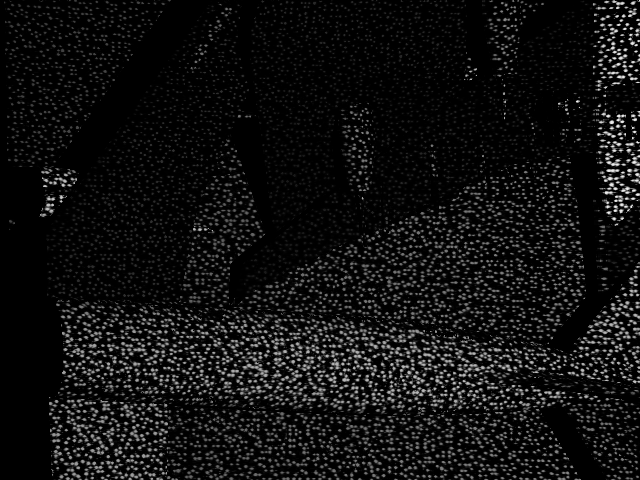}\\
            \scriptsize{Input for Connecting}
         \end{minipage}
        \begin{minipage}[t]{0.32\hsize}
          \centering
            \includegraphics[width=\hsize]{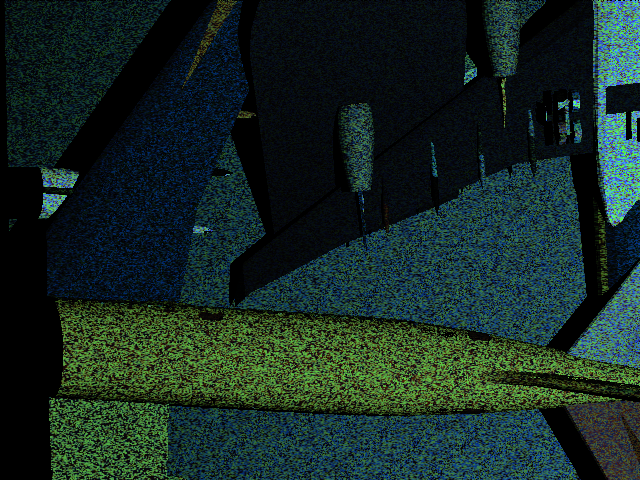}\\
            \scriptsize{Input for ours}
         \end{minipage}\\
        \footnotesize{(a) Input images for each method.}
    \end{minipage}\\
        \vspace{0.7em}
    \begin{minipage}{\hsize}
    \centering
      \includegraphics[width=\hsize]{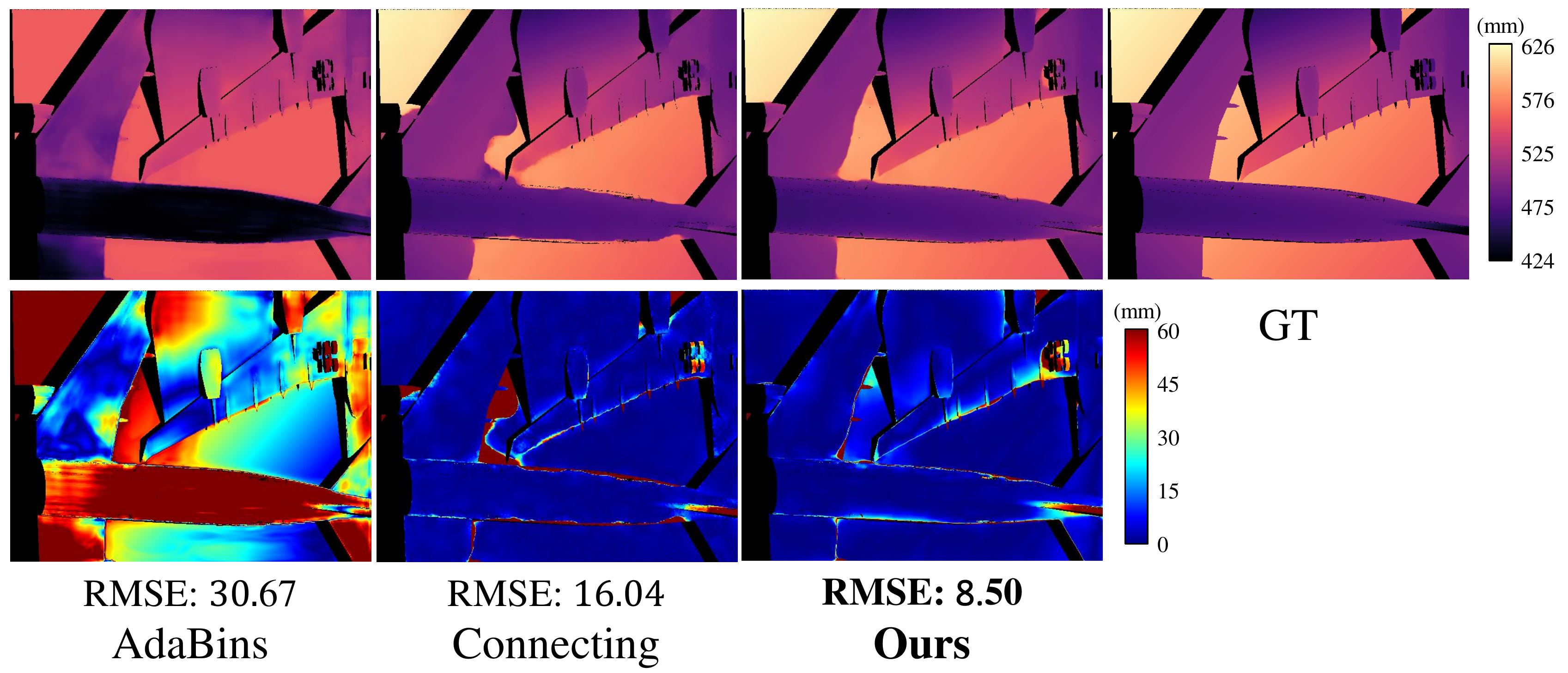}\\
      \footnotesize{(b) Depth results. Top: Estimated depth maps. Bottom: Visualized errors.}
    \end{minipage}\\
    \vspace{0.7em}
    \begin{minipage}{\hsize}
    \centering
      \includegraphics[width=\hsize]{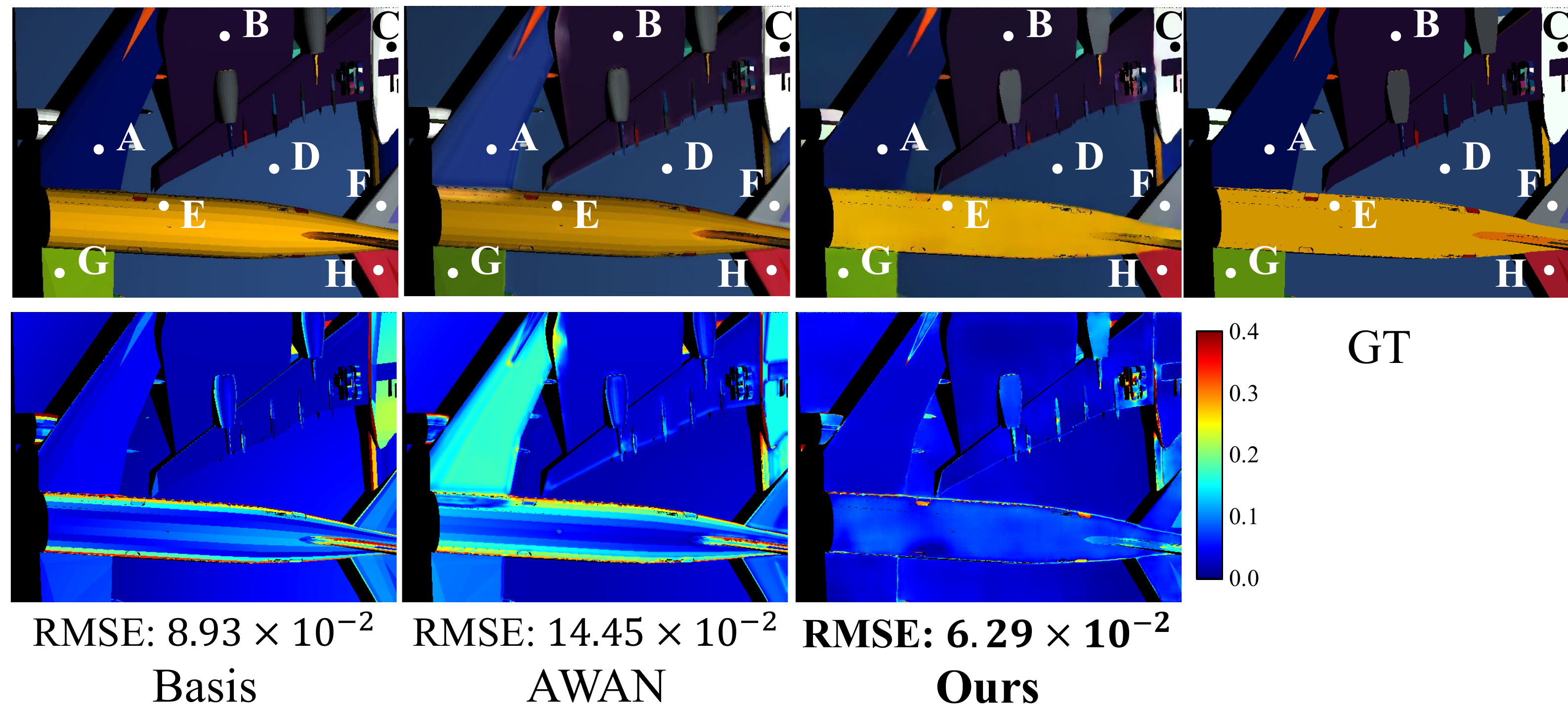}\\
      
    \footnotesize{(c) Spectral reflectance results. Top: sRGB color representation converted from the estimated spectral reflectances. Bottom: Visualized errors.}
    \end{minipage}\\
    \vspace{0.7em}
    \begin{minipage}[t]{0.24\hsize}
    \centering
      \includegraphics[width=\hsize]{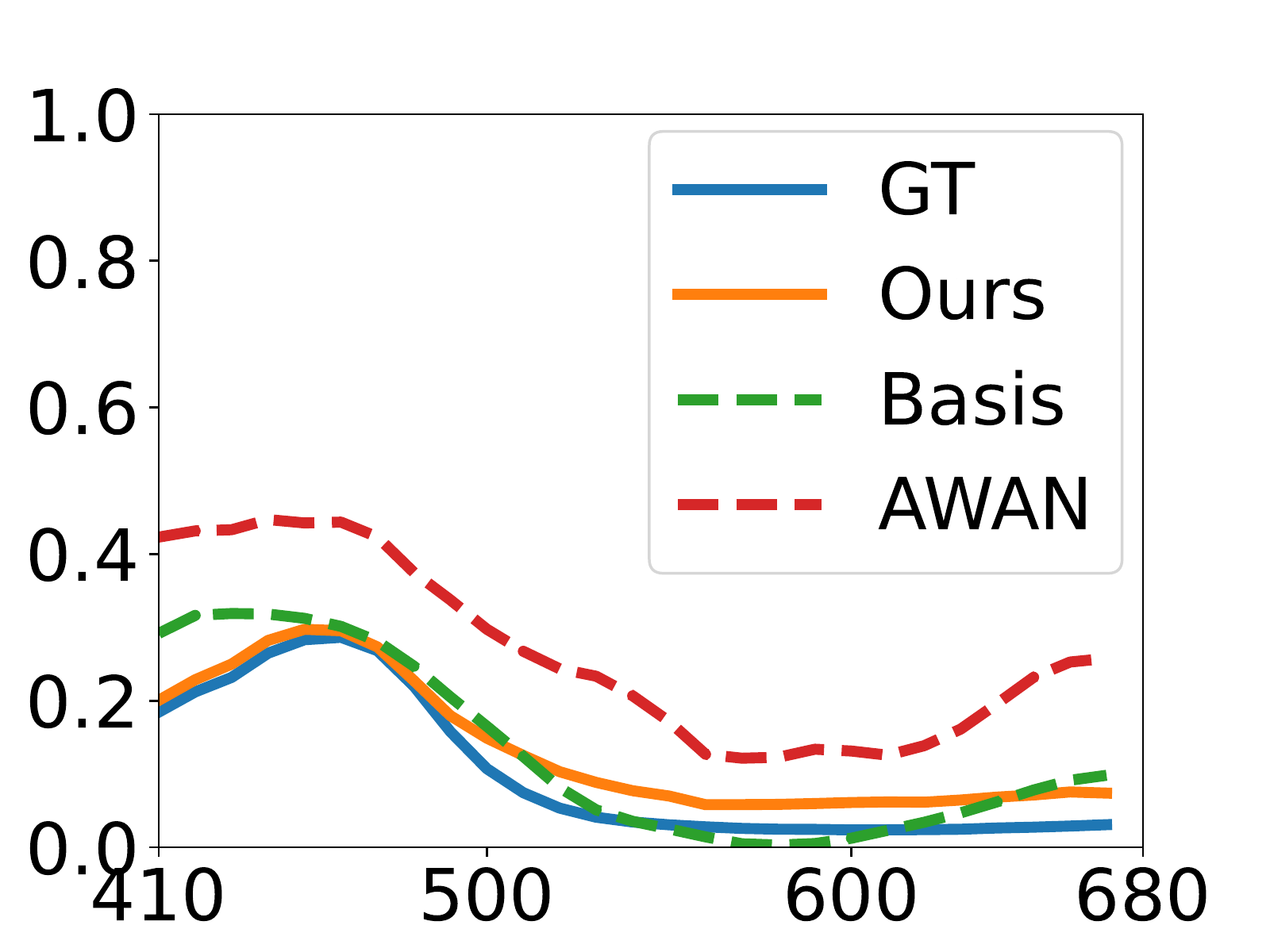}
            \scriptsize{A}
    \end{minipage}
    \begin{minipage}[t]{0.24\hsize}
    \centering
      \includegraphics[width=\hsize]{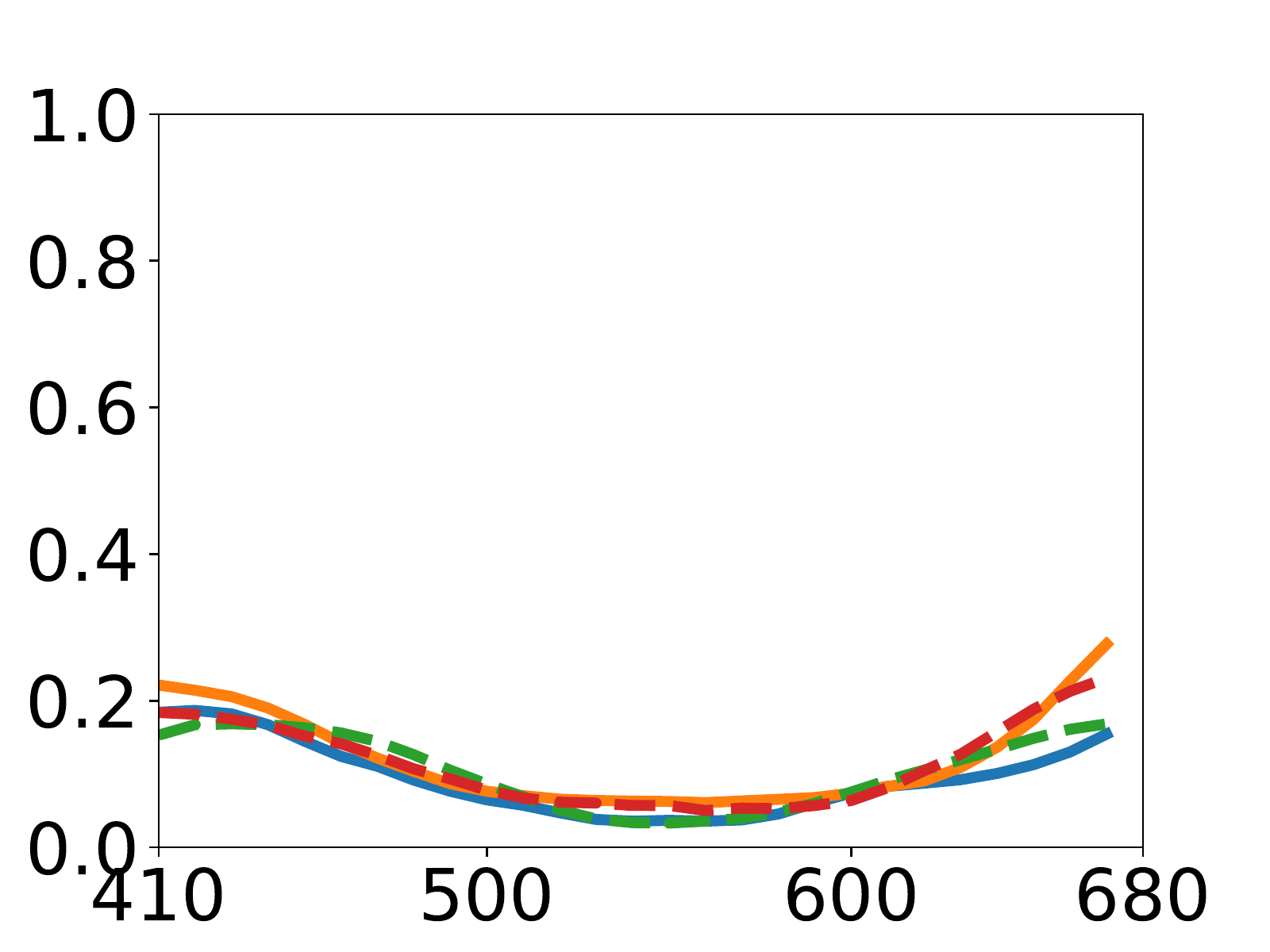}
            \scriptsize{B}
    \end{minipage}
    \begin{minipage}[t]{0.24\hsize}
    \centering
      \includegraphics[width=\hsize]{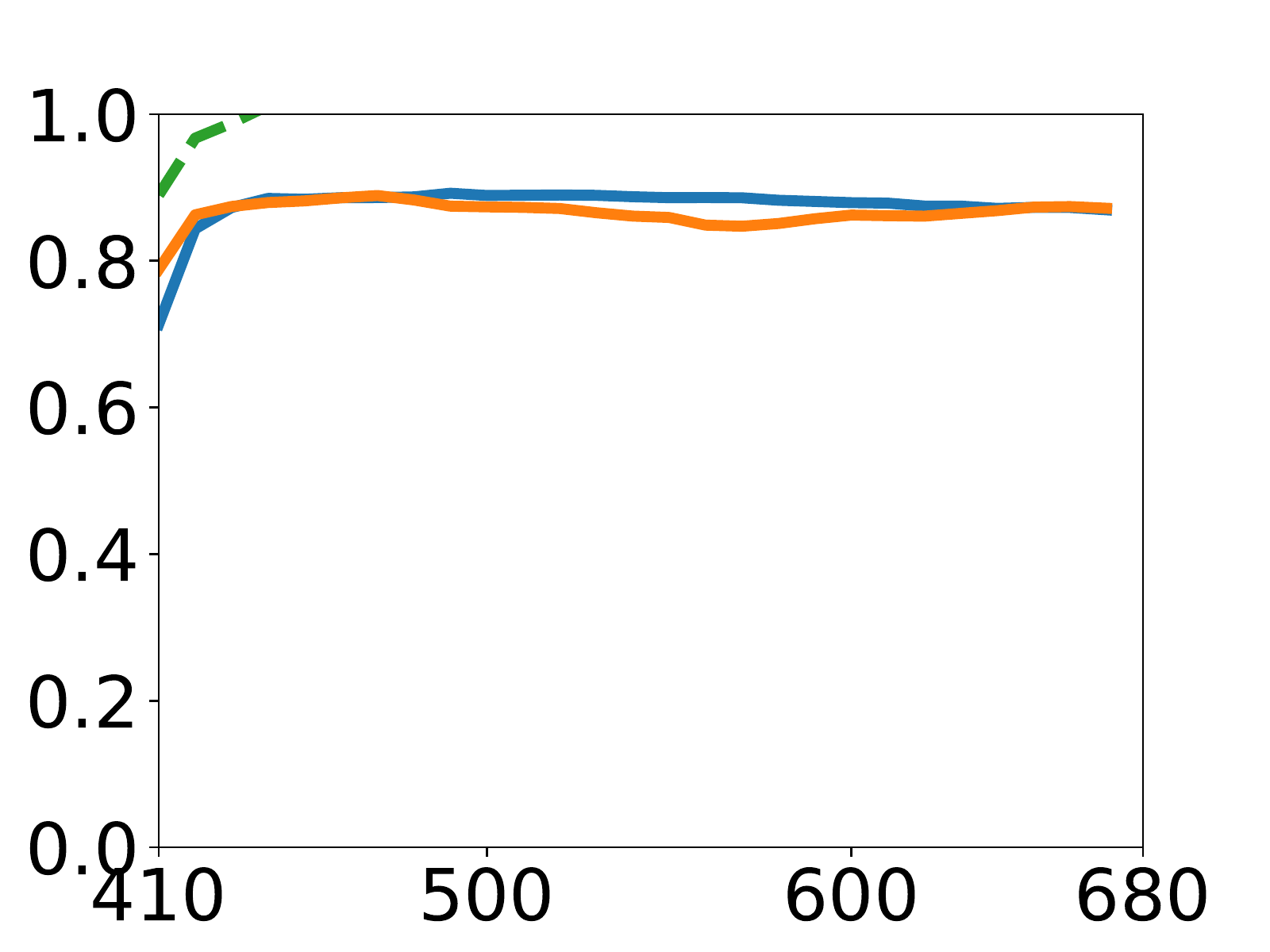}
            \scriptsize{C}
    \end{minipage}
    \begin{minipage}[t]{0.24\hsize}
    \centering
      \includegraphics[width=\hsize]{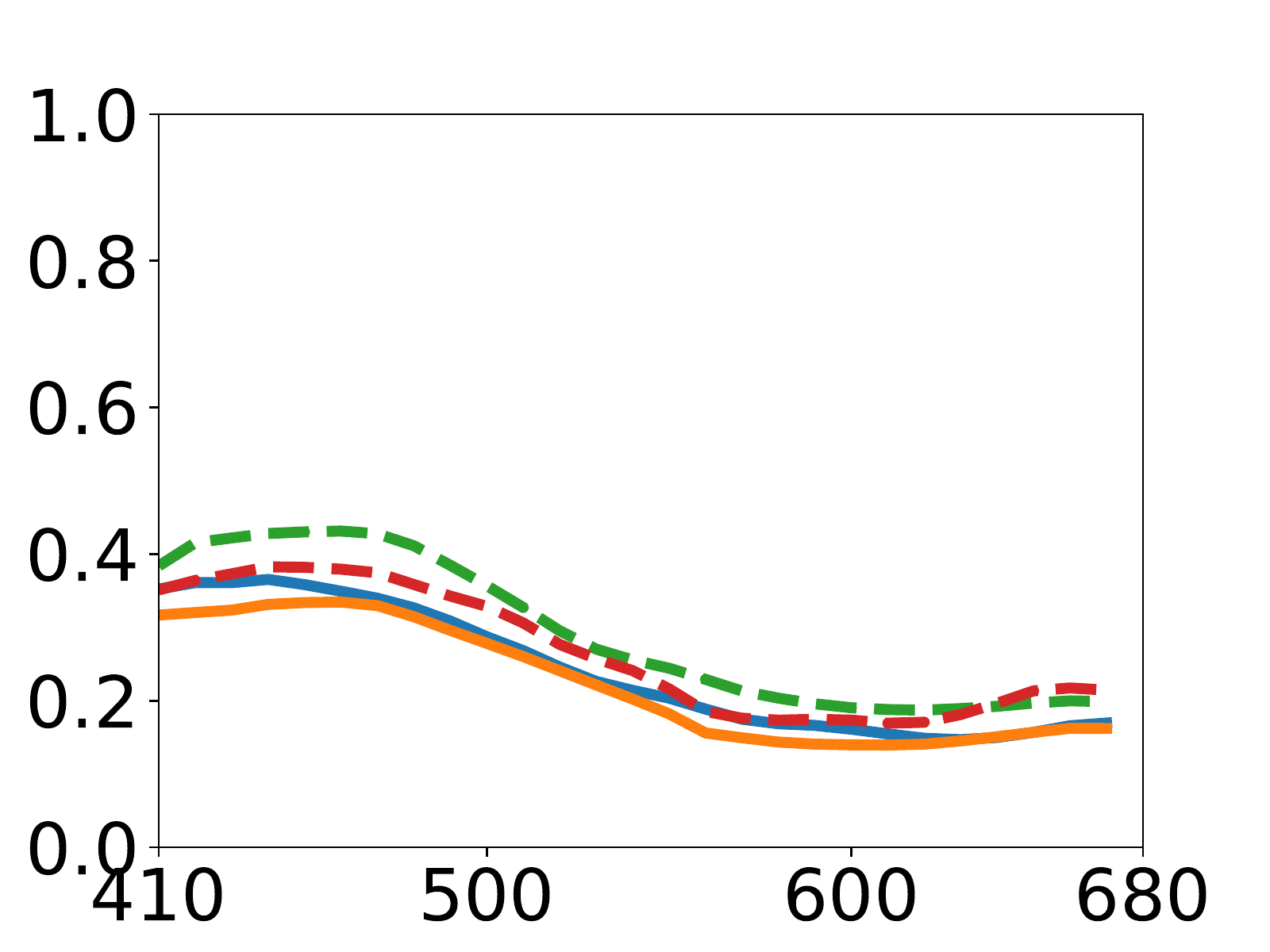}
            \scriptsize{D}
    \end{minipage}\\
    \begin{minipage}[t]{0.24\hsize}
    \centering
      \includegraphics[width=\hsize]{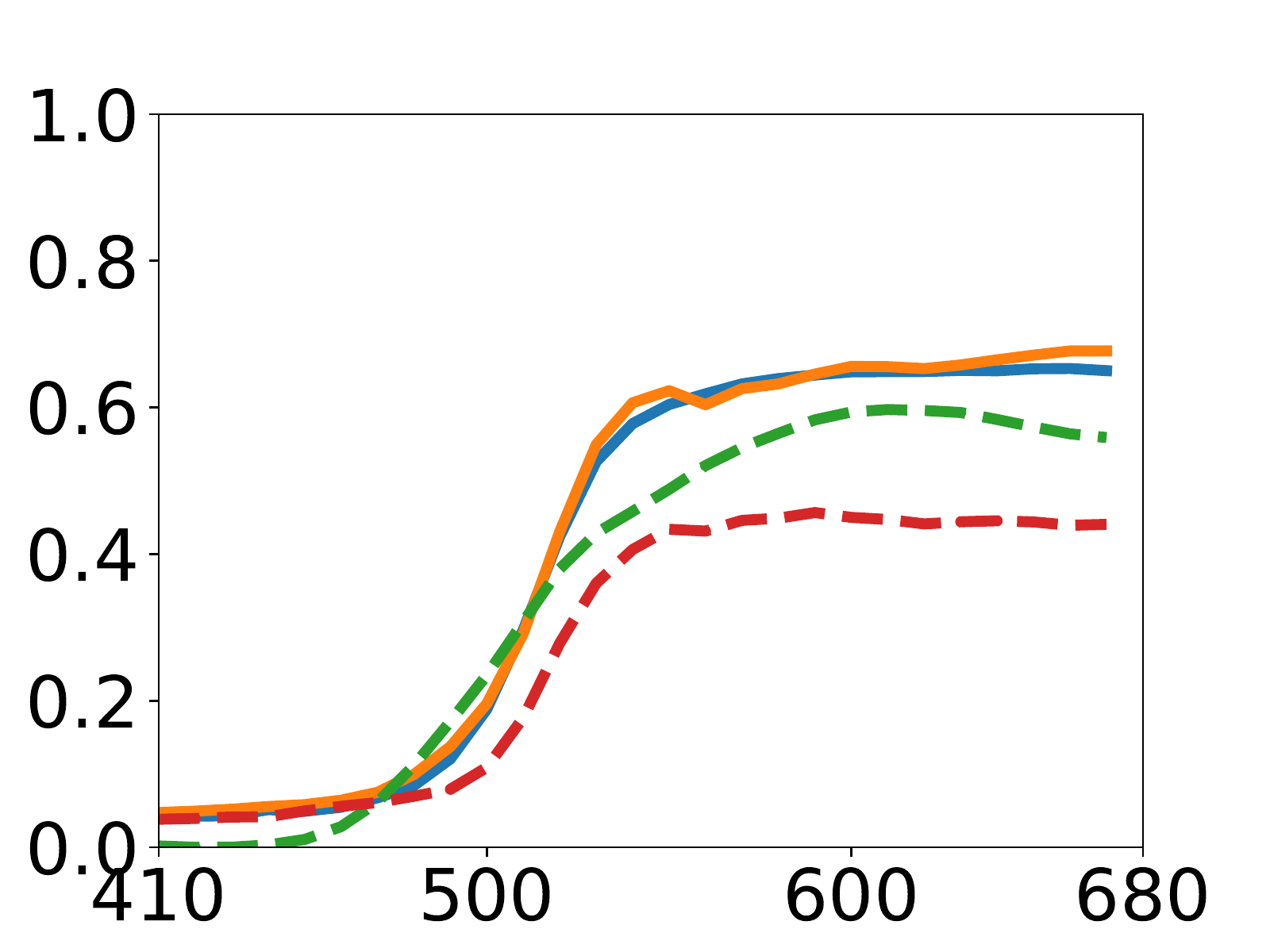}
            \scriptsize{E}
    \end{minipage}
    \begin{minipage}[t]{0.24\hsize}
    \centering
      \includegraphics[width=\hsize]{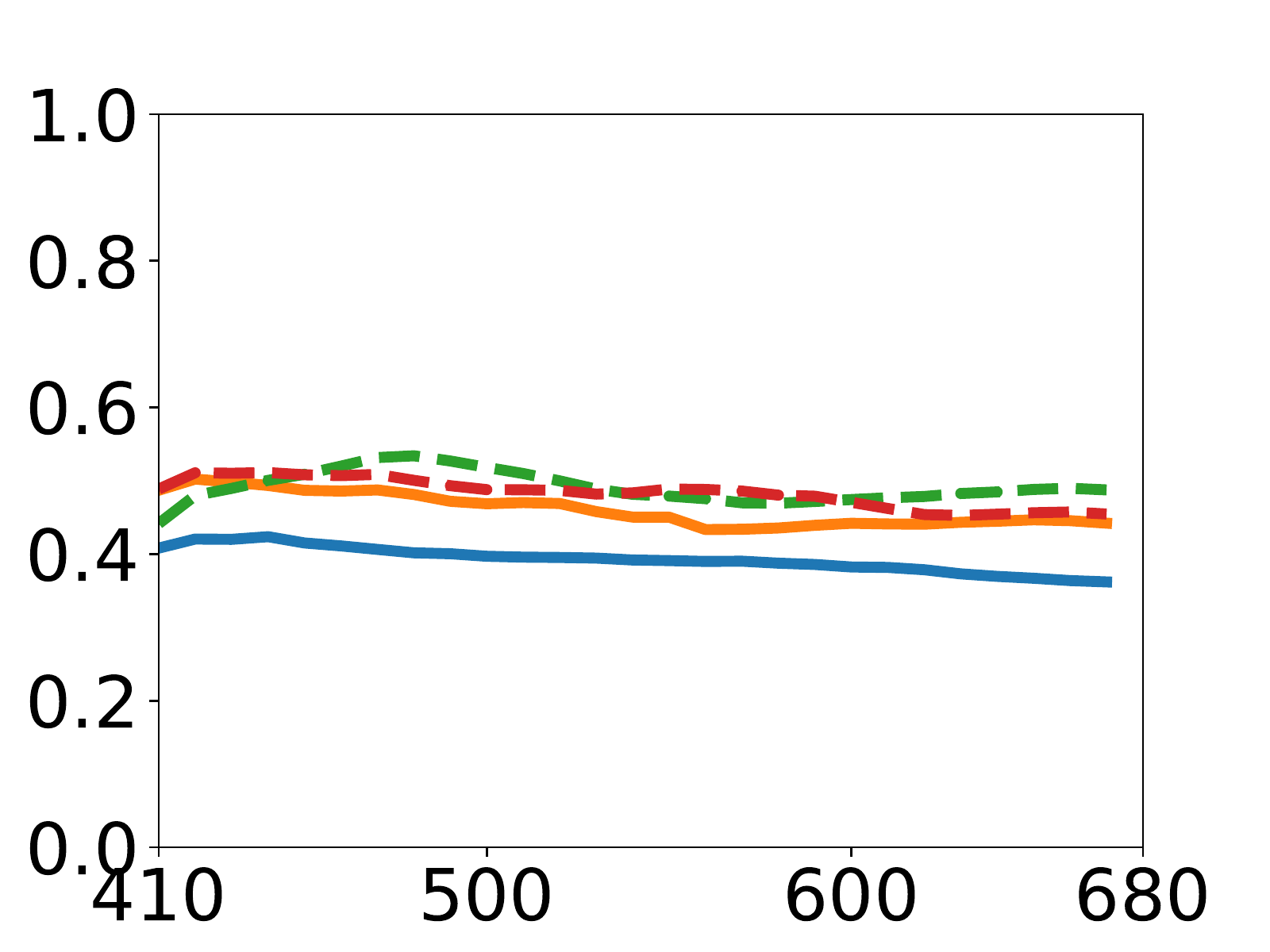}
            \scriptsize{F}
    \end{minipage}
    \begin{minipage}[t]{0.24\hsize}
    \centering
      \includegraphics[width=\hsize]{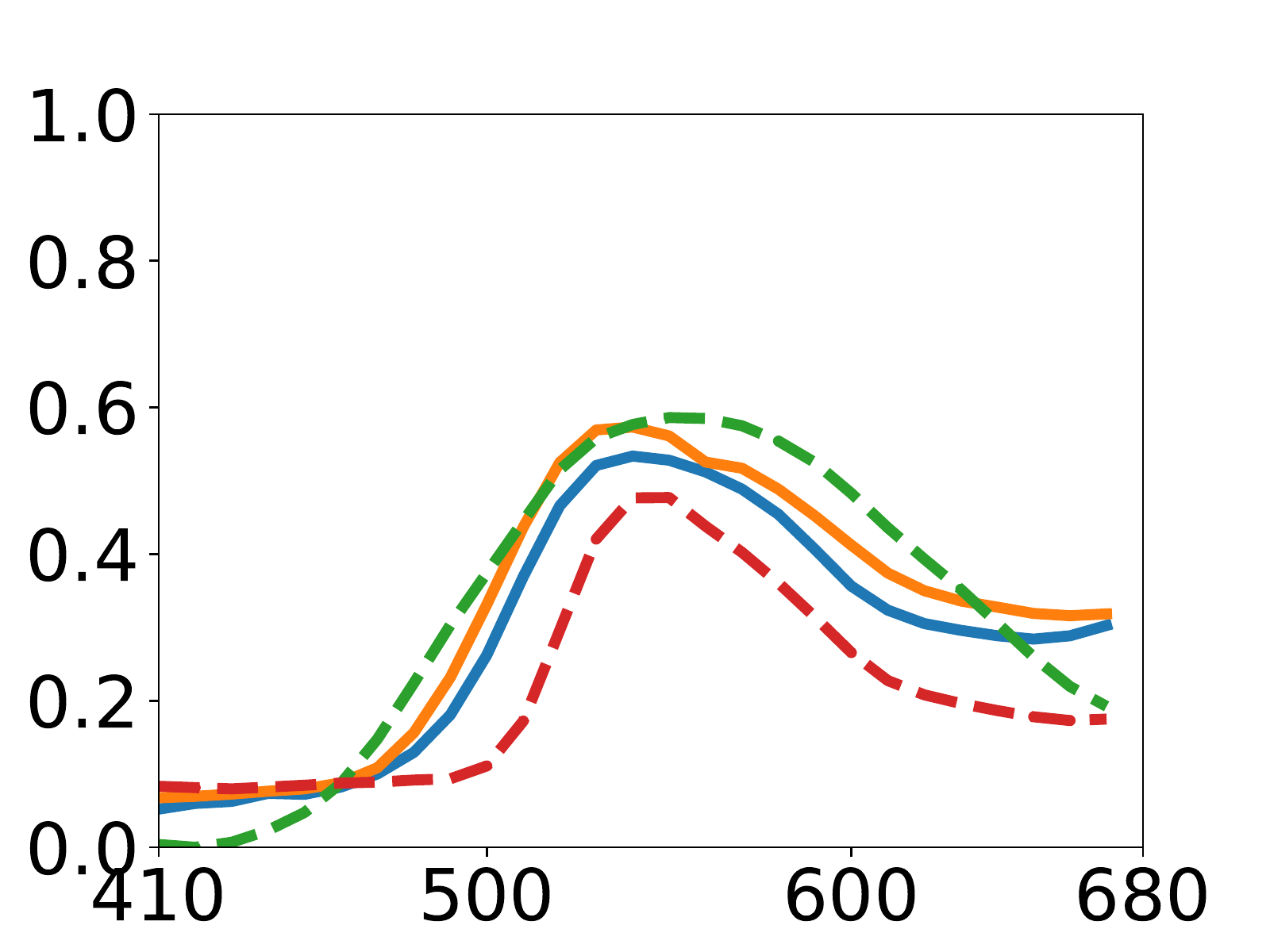}
            \scriptsize{G}
    \end{minipage}
    \begin{minipage}[t]{0.24\hsize}
    \centering
      \includegraphics[width=\hsize]{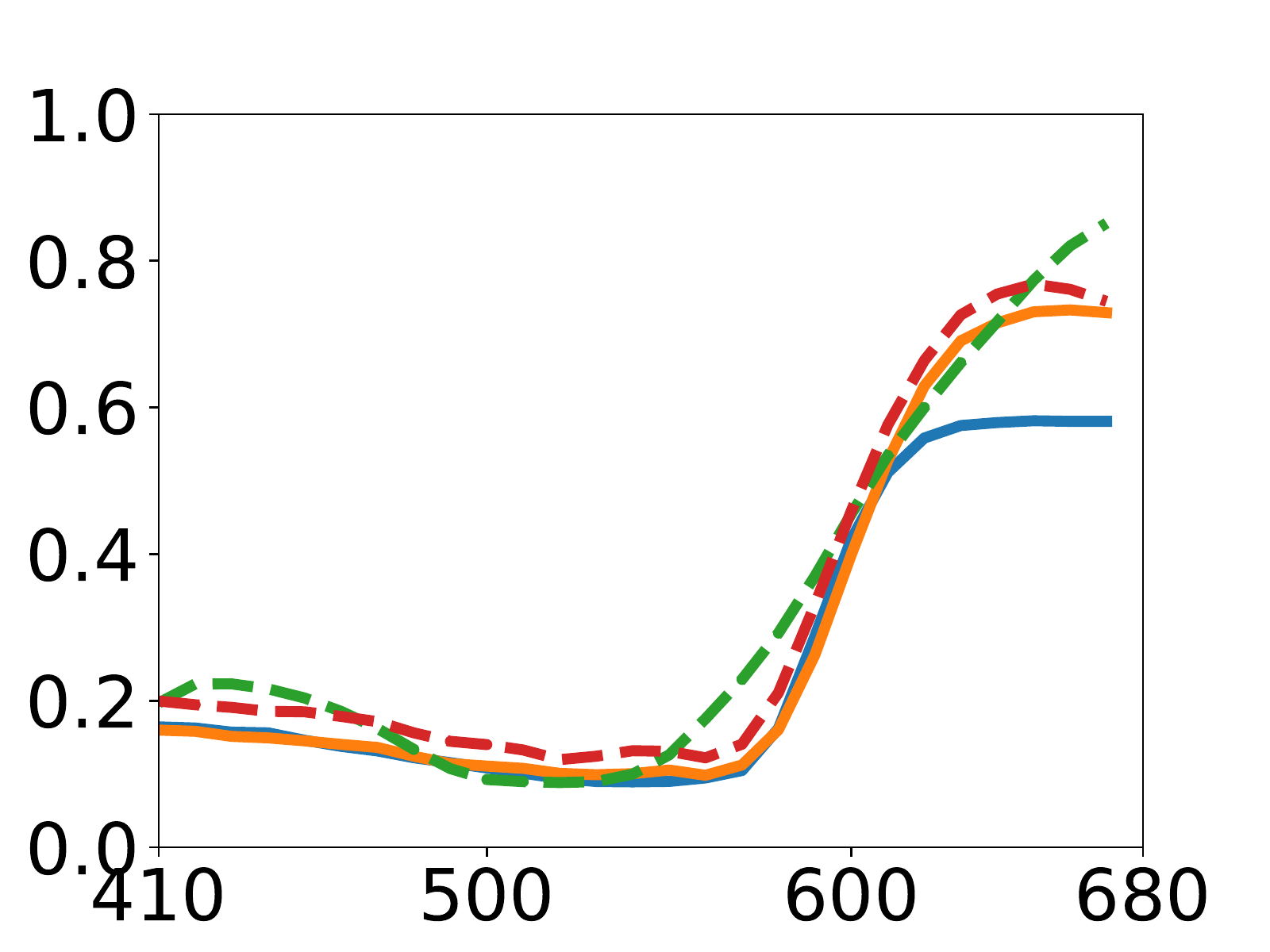}
            \scriptsize{H}
    \end{minipage}\\\vspace{0.4em}
      \footnotesize{(d) Estimated spectral reflectance results on 8 sample points.}
      \vspace{-1mm}
    \caption{Synthetic comparison with the state-of-the-art methods.}
    \label{fig:cgresult}
      \vspace{-2mm}
\end{figure}

\vspace{-4mm}
\subsubsection{Comparison with State-of-the-Art Methods}

Since there is no existing single-shot hyperspectral-depth reconstruction method directly applicable to our setup, we compare our method with state-of-the-art single-shot depth reconstruction methods and single-shot spectral reconstruction methods, respectively.
For the depth evaluation, we compare our method with state-of-the-art AdaBins~\cite{bhat2021adabins}, which learns the depth from a standard RGB image without any dot pattern, and Connecting the Dots~\cite{riegler2019connecting}, which learns the depth from a single gray-scale image with Kinect dot pattern.
Their networks were retrained using our dataset. Since their input images are different from ours, we re-rendered RGB images without the dot pattern (under white illumination) and gray-scale Kinect dot pattern images, respectively. The examples of the input images are shown in Fig.~\ref{fig:cgresult}(a).
For the spectral reflectance evaluation, we compare our method with two spectral reconstruction methods from a single RGB image: A widely-applied method using spectral reflectance basis functions (Basis)~\cite{Han2} and a state-of-the-art deep learning-based AWAN~\cite{li2020adaptive}.
We used the RGB image without the dot pattern (under white illumination) as their inputs, which is the same input as AdaBins. For Basis, the spectral basis functions were calculated from 1,269 Munsell spectral reflectances of our training data. For AWAN, we retrained the model using our dataset.

Table~\ref{table:stateoftheart} summarizes the overall quantitative evaluation on all 256 test scenes. 
We can observe that our method yields the best results for both the depth and the spectral reflectance. In contrast to the compared methods that only focus on a single property, our method jointly reconstructs the depth and the spectral reflectance by training the network model using both geometric losses and photometric losses, leading the improved performance to each other. 

Figure~\ref{fig:cgresult}(b) provides the qualitative results of the depth reconstruction, where our method provides a more accurate depth map.
Figure~\ref{fig:cgresult}(c) shows the visual comparison of sRGBs (top row), which was converted from the estimated spectral reflectances, and the error maps for the estimated spectral reflectances (bottom row), where RMSE for all wavelengths is visualized for each pixel. We can confirm that our sRGB result is the closest to the ground truth and represents the object's inherent spectral reflectance less affected by the shading, compared with the sRGB results of the existing methods that do not consider the shading (depth) information.
Figure~\ref{fig:cgresult}(d) shows the spectral reflectance results on eight sample points. Our method can reconstruct accurate spectral reflectances representing correct spectral shapes as well as correct relative scales. This is because that our method can benefit from the color-dot projection to acquire nine-band information and depth information, while the existing single-shot spectral reconstruction methods only rely on a standard three-band RGB image.

\vspace{-4mm}
\subsubsection{Ablation Study}
\vspace{-2mm}

\noindent
\textbf{Loss comparison for disparity estimation:}
To confirm the contribution of the individual geometric loss of the disparity estimation network (disparity loss $\mathcal{L}_{\rm D}$, disparity edge loss $\mathcal{L}_{\rm DE}$, and pattern loss $\mathcal{L}_{\rm P}$), we trained the disparity estimation network only and compared different loss combinations. Table~\ref{table:depthloss} shows the depth reconstruction performance with the four loss combinations.
We can observe that, if we add the disparity edge loss $\mathcal{L}_{\rm DE}$, the depth accuracy is significantly improved especially for large errors evaluated in the metrics $\theta_3$, indicating that the edge loss contributes to reducing the boundary errors. The pattern loss $\mathcal{L}_{\rm P}$ significantly reduces overall RMSE, demonstrating the effectiveness of the color-dot pattern matching. We can also confirm that the best result can be achieved by using all the losses.

\vspace{2mm}
\noindent
\textbf{Effectiveness of joint training:}
To demonstrate the effectiveness of the joint training of the depth and the spectral reflectance, we compare our full model with the following network models. (i) Disparity estimation network, which applies only the disparity estimation network for the disparity training, as compared in the previous paragraph (i.e., the best result of Table~\ref{table:depthloss}). (ii) Spectral reconstruction network, which applies only the spectral reconstruction network using only the single color-dot image input. The networks (i) and (ii) mean the cases of the separated network training at our setup. (iii) Joint network training without the depth input, which applies both the disparity estimation network and the spectral reconstruction network, but does not apply the depth input for the spectral reconstruction network. (iv) Joint network training without the illumination input, which applies both the networks, but does not apply the warped illumination input for the spectral reconstruction network. The networks (iii) and (iv) are compared to confirm the importance of the depth and the illumination inputs for estimating the spectral reflectance.

\begin{table}[t]
\centering
 \caption{Depth accuracy comparison with respect to different loss combinations for the disparity estimation network.}
 \centering
  \vspace{-1mm}
  \small
  \begin{tabular}{lcccc}
   \toprule
    & $\theta_1$ $\uparrow$ & $\theta_2$ $\uparrow$ & $\theta_3$ $\uparrow$ & RMSE $\downarrow$  \\
   \midrule          
   $\mathcal{L}_{\rm D}$ &97.89 & 98.71 & 99.04 & 8.01 \\
   $\mathcal{L}_{\rm D}+\mathcal{L}_{\rm DE}$ & 97.93 & 98.89 & 99.32 & 7.38 \\
   $\mathcal{L}_{\rm D}+\mathcal{L}_{\rm P}$ & 98.00 & 98.90 & 99.10& 7.22 \\
   $\mathcal{L}_{\rm D}+\mathcal{L}_{\rm DE}+\mathcal{L}_{\rm P}$ & \textbf{98.03} & \textbf{99.12} & \textbf{99.38} & \textbf{6.80} \\
   \bottomrule
  \end{tabular}
 \label{table:depthloss}
\end{table}

\begin{table}[t]
 \caption{Effectiveness of joint training.}
 \centering
  \vspace{-1mm}
  \small
  \begin{tabular}{l|l|cc}
   \hline 
    \multicolumn{2}{l|}{} & \makecell[c]{Depth\\RMSE $\downarrow$} & \makecell[c]{Reflectance\\RMSE $\downarrow$\\($\times10^{-2}$)}\\
   \hline          
   \multicolumn{2}{l|}{Disparity estimation network}  & 6.80 & - \\
   \hline        
   \multicolumn{2}{l|}{Spectral reconstruction network} & - & 5.79  \\
   \hline        
   \multirow{3}{*}{Joint} & w/o depth input & 6.24 & 5.69 \\
\cline{2-4}
   & w/o illumination input & 6.32 & 5.75 \\
\cline{2-4}    
   & full model & \textbf{6.10} & \textbf{5.30} \\
   \hline 
  \end{tabular}
 \label{table:ablation}
\end{table}

Table~\ref{table:ablation} shows the results of the comparison. From the results, we can observe that joint training certainly provides better performance compared with the separated training of the depth and the spectral reflectance. In addition, both the depth and the warped illumination inputs to the spectral reconstruction network contribute to the performance improvement for estimating object-inherent (shading- and illumination-irrelevant, in other words) spectral reflectance. Interestingly, the depth result also can be significantly improved by using the warped illumination input, because the illumination spectrum pattern corresponding to the input image is accurate only when the disparity is correct, suggesting that the errors of the spectral reflectances can be back-propagated to update the disparity estimation network.

\begin{figure}[!tbp]
  \centering
 \includegraphics[width=\hsize]{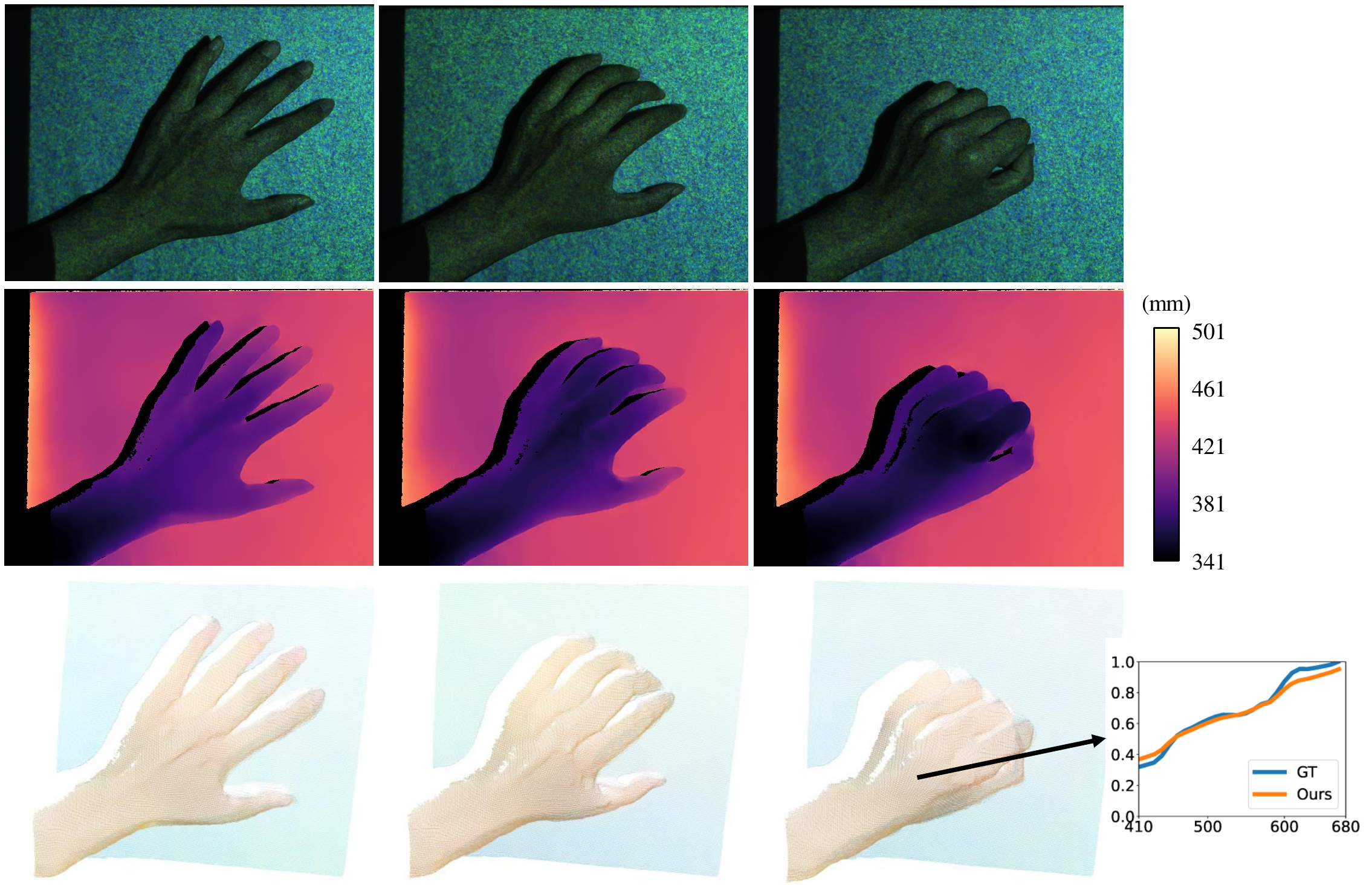}
  \caption{The reconstruction results for a dynamic scene (hand). Top: The sequentially captured input color-dot images. Middle: The estimated depth maps from each input image. Bottom: The 3D point clouds converted from each estimated depth map and the spectral reflectance result for one sample point (rightmost). Each 3D point is colored by sRGB, which is converted from the corresponding estimated spectral reflectance.}
  \label{fig:hand}
       \vspace{-2mm}
\end{figure}

\subsection{Results on Real Scenes}
\label{ssec:realscenes}
We next evaluate our method for real scenes.
Because our method can realize the single-shot reconstruction of the depth and the spectral reflectance, we applied our method to a dynamic scene with a moving hand using a successive capturing mode of the camera. Figure~\ref{fig:hand} shows the captured input color-dot images (top row), the estimated depth maps from each input image (middle row) and the 3D point clouds converted from each depth map (bottom row).
Each 3D point of the point cloud is colored by sRGB, which is converted from the corresponding estimated spectral reflectance. We also show the spectral reflectance result for one sample point in the right-bottom figure. From the results, we can confirm that our method performs well in the dynamic real scene.

\begin{figure}[!tbp]
  \centering
 \includegraphics[width=\hsize]{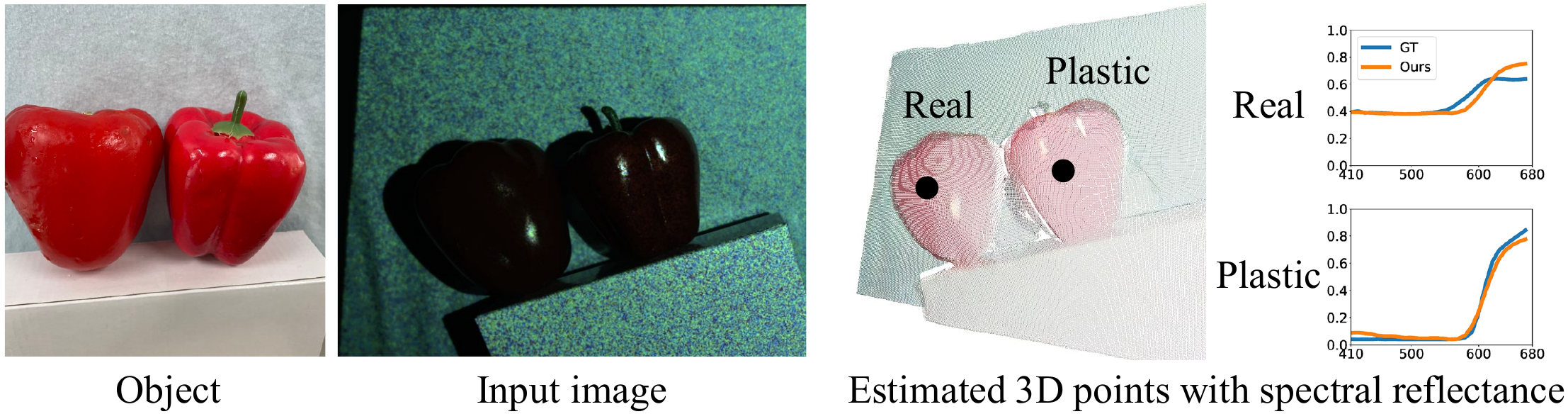}
 \footnotesize{(a) Red bell pepper}
 \includegraphics[width=\hsize]{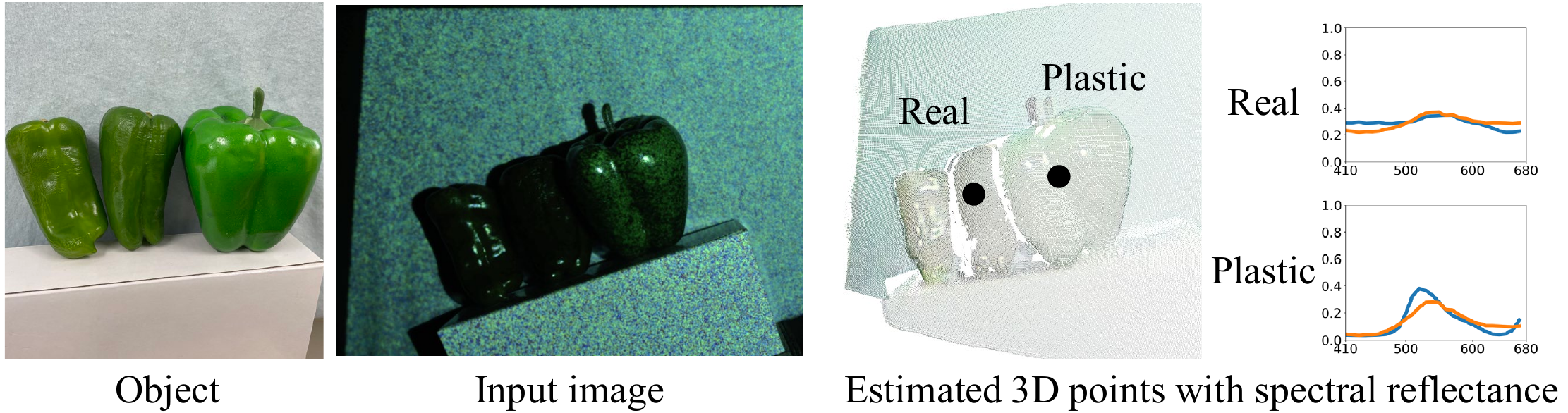}
 \footnotesize{(b) Green bell pepper}
  \caption{Which is real? We captured real (a) red and (b) green bell peppers, as well as plastic models, to show an example application for material discrimination. From left to right, we show the object images for reference, the input images to our system, the resultant 3D point clouds with sRGB color representation, and the estimated spectral reflectances for sampled points.}
  \label{fig:real}
       \vspace{-2mm}
\end{figure}

One important application of the spectral reflectance reconstruction is to differentiate the materials that have similar colors, but different spectral reflectances, because the spectral reflectances provide much richer information than the RGB tristimulus values. To demonstrate this, we captured real and plastic bell peppers. As the objects shown in the left column of Fig.~\ref{fig:real}, it is hard to differentiate the real and the plastic bell peppers only from the color appearance. In contrast, we can confirm the difference of the spectral reflectances from our spectral reconstruction results, as shown in the right column. Beyond that, our system can reconstruct dense 3D points using the estimated depth maps. 
Additional results on real scenes reconstructed by our proposed system can be seen in the supplemental video.

\subsection{Discussion and Limitations}
To offer a theoretical insight on the spectral reflectance estimation, we conducted a condition-number analysis on the system matrix consisting of the products of the projector's RGB illumination spectrums and RGB camera sensitivities. The condition number is 1275.3, which indicates that the direct linear inverse problem is highly ill-posed. As commonly performed~\cite{Park,Han2}, if we introduce a spectral basis model (e.g., 8 bases) and a smoothness constraint to the spectral reflectance, the condition number reduces to 13.0, which means that the problem is solvable. Although we solved the ill-posed problem without such constraints by exploiting deep-learning-based reconstruction, we consider that bridging the theoretical analysis and the learning-based method could be one of the important future directions.


Our method has several limitations. First, our method will degrade the performance under the existence of strong ambient illumination because it makes the color-dot extraction by LCN more difficult and it also changes the illumination spectrum of the color dot. Second, heavy occlusions will lead to a large area of cast shadow, resulting in a highly incomplete depth map. Third, similar to other structured-light methods, our method is difficult to reconstruct the dark objects that do not reflect the projector light sufficiently.

\section{Conclusion}
In this paper, we have proposed a novel single-shot system to simultaneously acquire scene depth and spectral reflectance using a standard RGB camera and an off-the-shelf projector.
Our system utilizes a single color-dot projection to simultaneously provide geometric and spectral observations.
To effectively reconstruct the depth and the spectral reflectance in a joint training manner, we have built an end-to-end deep neural network architecture by incorporating a geometric color-dot pattern loss and a photometric spectral reflectance loss. Experimental results using both synthetic and real-world data have demonstrated the potential of our system for a high-fidelity 3D sensing technology.
Our dataset and spectral renderer for the dataset generation are available in our project page (\url{http://www.ok.sc.e.titech.ac.jp/res/DHD/}).

\vspace{2mm}
\noindent \textbf{Acknowledgment}
This work was partly supported by JSPS KAKENHI Grant Number 17H00744 and 21K17762. We thank Tatsuhiko Tezuka for assisting in our experiments.

{\small
\bibliographystyle{ieee_fullname}
\bibliography{egbib}
}

\end{document}


\title{Supplementary Material:\\Deep Hyperspectral-Depth Reconstruction Using Single Color-Dot Projection}

\author{Chunyu Li, Yusuke Monno, and Masatoshi Okutomi\\
Tokyo Institute of Technology, Tokyo, Japan\\
{\tt\small \{lchunyu,ymonno\}@ok.sc.e.titech.ac.jp, mxo@ctrl.titech.ac.jp}
}
\maketitle

\section*{Supplementary Material: Network Architecture}

\begin{figure*}[b]
  \centering
 \includegraphics[width=\hsize]{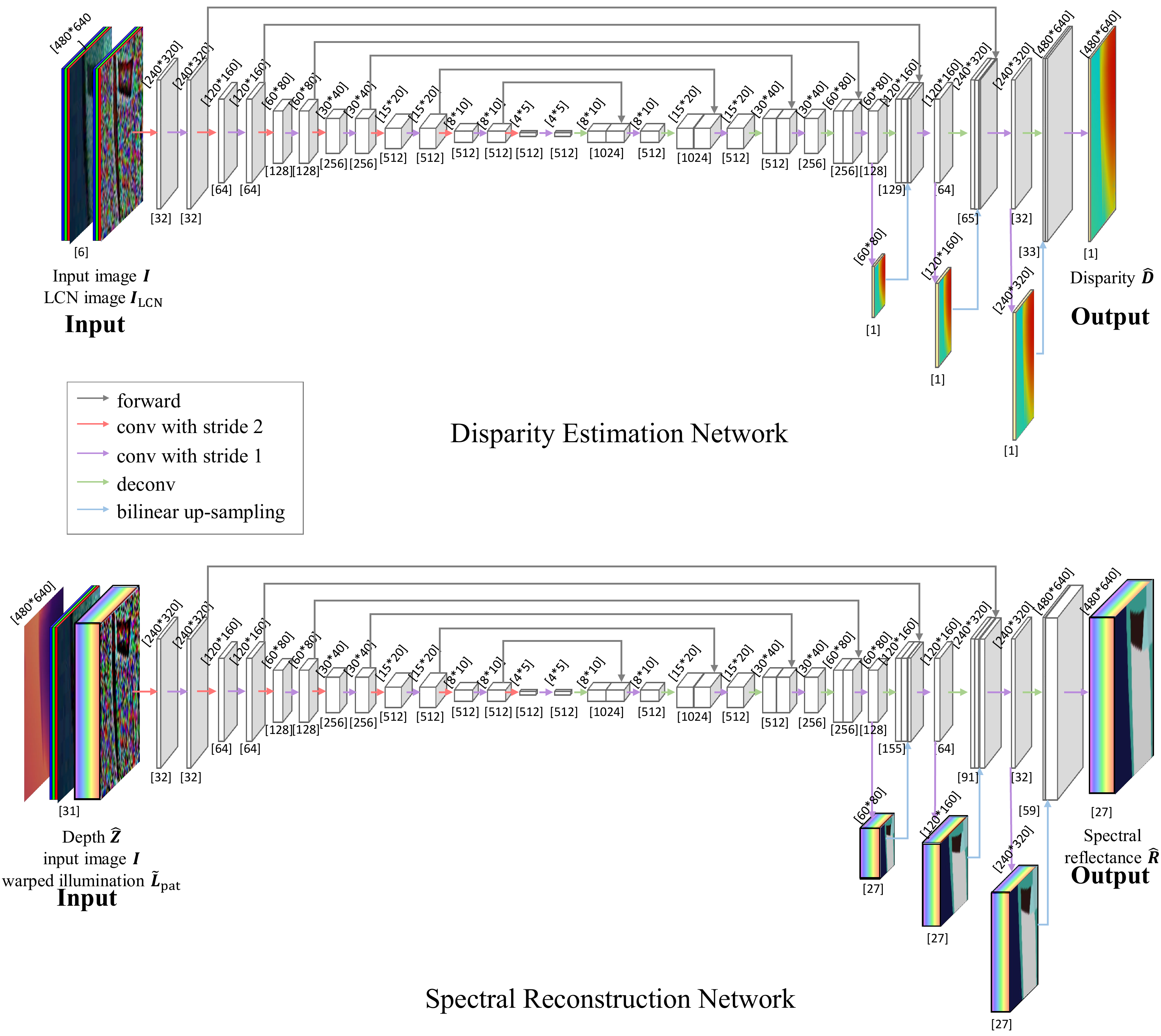}
  \caption{Detailed architectures of the disparity estimation network (top) and the spectral reconstruction network (bottom).}
  \label{fig:architecture}
      \vspace{-5mm}
\end{figure*}

We designed the architectures of the disparity estimation network and the spectral reconstruction network based on the Disparity Decoder presented in the previous work~\cite{riegler2019connecting}. The detailed architectures are shown in Fig.~\ref{fig:architecture} of the next page.
Each network consists of a contractive part and an expanding part with long-range links between them. 
The contracting part contains convolution layers with the strides of 2, resulting in a total downsampling factor of 128. The expanding part of the network then gradually and nonlinearly upsamples the feature maps by also taking into account the features from the contractive part. Convolutional filter sizes decrease towards deeper layers of the network: 7$\times$7 for the first and second layers, 5$\times$5 for the following two layers and 3$\times$3 for the layers after the fifth layer.
In total, each network has 32 convolution layers and each of them is followed by ReLU. The final layer is followed by a scaled sigmoid non-linearity which constrains the output to the range between 0 and the maximum.

{\small
\bibliographystyle{ieee_fullname}
\bibliography{egbib}
}